\documentclass[sigconf]{acmart}

\AtBeginDocument{%
  \providecommand\BibTeX{{%
    \normalfont B\kern-0.5em{\scshape i\kern-0.25em b}\kern-0.8em\TeX}}}

\setcopyright{acmlicensed}
\copyrightyear{2025}
\acmYear{2025}
\acmDOI{XXXXXXX.XXXXXXX}

\acmConference[KDD '25]{ KDD '25: ACM SIGKDD Conference on Knowledge Discovery and Data Mining}{August 03--07,
  2025}{Toronto, ON}
\acmISBN{978-1-4503-XXXX-X/24/06}

\usepackage{multirow}
\usepackage{subfig}

\usepackage{algorithm}
\usepackage{algorithmic}

\usepackage{tikz}
\usetikzlibrary{arrows.meta,positioning}

\begin{document}





\title{Learning for Long-Horizon Planning via Neuro-Symbolic Abductive Imitation}
\author{Jie-Jing Shao}
\authornote{Both authors contributed equally to this research.}
\affiliation{%
  \institution{National Key Laboratory for Novel Software Technology}
  \country{Nanjing University, Nanjing, China}
}
\email{shaojj@lamda.nju.edu.cn}

\author{Hao-Ran Hao}
\authornotemark[1]
\affiliation{%
  \institution{National Key Laboratory for Novel Software Technology}
  \country{Nanjing University, Nanjing, China}
}
\email{hrhao.ai@gmail.com}

\author{Xiao-Wen Yang}
\affiliation{%
  \institution{National Key Laboratory for Novel Software Technology\\School of Artificial Intelligence}
  \country{Nanjing University, Nanjing, China}
}
\email{yangxw@lamda.nju.edu.cn}

\author{Yu-Feng Li}
\authornote{Corresponding Author.}
\affiliation{%
  \institution{National Key Laboratory for Novel Software Technology\\School of Artificial Intelligence}
  \country{Nanjing University, Nanjing, China}
}
\email{liyf@nju.edu.cn}




\renewcommand{\shortauthors}{Jie-Jing Shao, Hao-Ran Hao, Xiao-Wen Yang and Yu-Feng Li.}

\begin{abstract}
Recent learning-to-imitation methods have shown promising results in planning via imitating within the observation-action space. However, their ability in open environments remains constrained, particularly in long-horizon tasks. In contrast, traditional symbolic planning excels in long-horizon tasks through logical reasoning over human-defined symbolic spaces but struggles to handle observations beyond symbolic states, such as high-dimensional visual inputs encountered in real-world scenarios. In this work, we draw inspiration from abductive learning and introduce a novel framework \textbf{AB}ductive \textbf{I}mitation \textbf{L}earning (ABIL) that integrates the benefits of data-driven learning and symbolic-based reasoning, enabling long-horizon planning. Specifically, we employ abductive reasoning to understand the demonstrations in symbolic space and design the principles of sequential consistency to resolve the conflicts between perception and reasoning. 
ABIL generates predicate candidates to facilitate the perception from raw observations to symbolic space without laborious predicate annotations, providing a groundwork for symbolic planning. 
With the symbolic understanding, we further develop a policy ensemble whose base policies are built with different logical objectives and managed through symbolic reasoning. Experiments show that our proposal successfully understands the observations with the task-relevant symbolics to assist the imitation learning. Importantly, ABIL demonstrates significantly improved data efficiency and generalization across various long-horizon tasks, highlighting it as a promising solution for long-horizon planning. Project
website: \url{https://www.lamda.nju.edu.cn/shaojj/KDD25_ABIL/}. 
\end{abstract}




\maketitle

\section{Introduction}

A long-standing goal in AI is to build agents that are flexible and general, able to accomplish a diverse set of tasks in open and novel environments, such as home robots for cooking meals or assembling furniture. 
These tasks generally require the agents to execute sequential decision-making, which is often formulated as a planning problem. Recently, the learning-based method, \textit{Imitation Learning}, has achieved remarkable success via imitating expert demonstrations, in a variety of domains, such as robotic manipulation~\cite{DBLP:journals/ijira/FangJGXWS19, MimicPlay}, autonomous driving~\cite{DBLP:journals/tits/MeroYDM22, DBLP:journals/tits/BhattacharyyaWPKMSK23} and language models~\cite{whitehurst1975language, DBLP:conf/nips/BrownMRSKDNSSAA20}. However, the theoretical studies~\cite{DBLP:conf/nips/RajaramanYJR20, DBLP:journals/pami/XuLY22} reveal that imitation learning can suffer from serious performance degradation due to the covariate shift between limited expert demonstrations and the state distribution actually encountered by the agents, especially in long-horizon tasks. 
In traditional AI literature, symbolic planners effectively generalize in long-horizon decision-making, via logical reasoning on the human-defined symbolic spaces~\cite{FikesN71, Gerevini20, DBLP:journals/jair/FoxL03}. 
However, they often simplify the perception process by relying on ground-truth symbols. 
Given observations and actions, pure logic-based methods struggle to map raw observations to human-defined symbolic spaces without predicate-level supervision. 

\begin{figure*}[th]
  \centering
  \includegraphics[width=.98\linewidth]{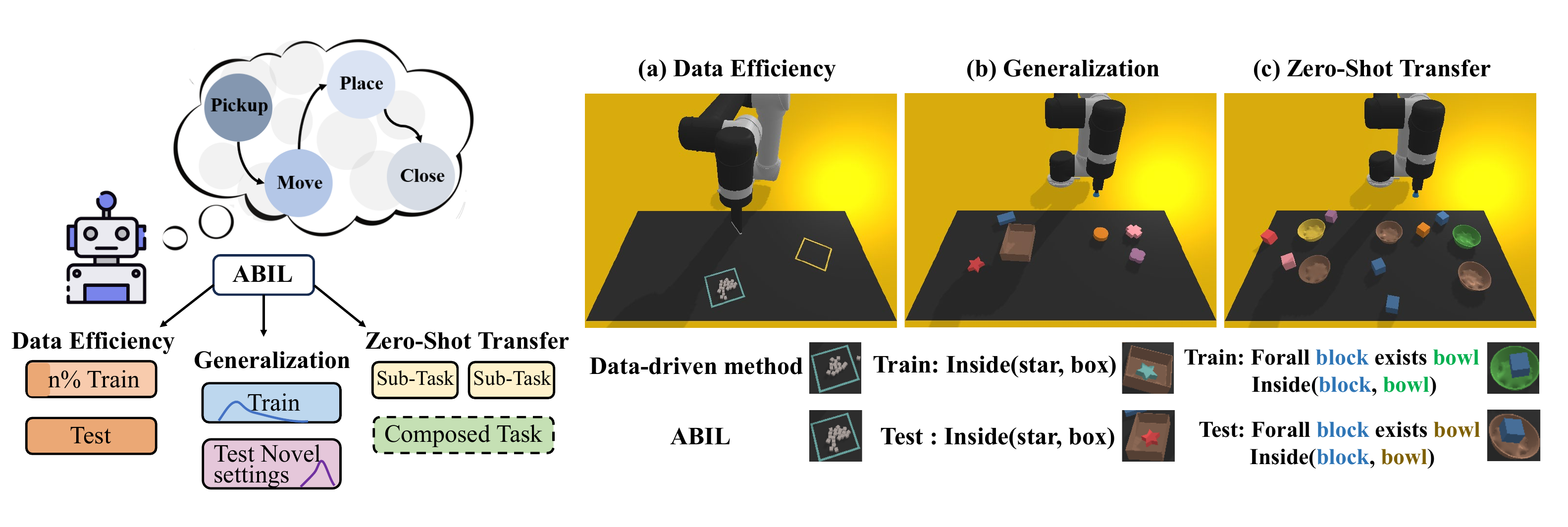}
  \vspace {-1em}
  \caption{Our framework, ABductive Imitation Learning (ABIL), achieves neuro-symbolic grounding of imitation in complex scenes while showing state-of-the-art results in data efficiency, generalization, and zero-shot transfer.}
  \label{ABIL_motivation}
\end{figure*}

To address these issues, efforts are underway to merge the advantages of learning-based and reasoning-based approaches into neuro-symbolic planning. 
\citeauthor{RPN}~\cite{RPN} propose the regression planning network, learning to predict symbolic sub-goals that need to be achieved before the final goals, thereby generating a long-term symbolic plan conditioned on high-dimensional observations. 
\citeauthor{DBLP:journals/jair/KonidarisKL18}~\cite{DBLP:journals/jair/KonidarisKL18} collect feasibility annotations and transition data under different symbolic operations to learn the symbolic representation of different observations. The learned representation enables traditional planning in the symbolic space and allows the acquisition of desired low-level controllers 
during inference. 
\citeauthor{DBLP:conf/iros/SilverCTKL21}~\cite{DBLP:conf/iros/SilverCTKL21} 
formalizes operator learning for neuro-symbolic planning, viewing operators as an abstraction model of the raw transition and generating the high-level plan skeletons. 
However, most of these positive results rely on the assumption that there are sufficient symbolic annotations to train the neural networks for mapping high-dimensional observations to symbolic states for logic-based planning, or there are prior low-level controllers to achieve the expected sub-goals perfectly. 
Compared to real-world applications, these approaches overlook the process of learning from demonstrations to imitate specific behaviors.
The most relevant work to ours is PDSketch~\cite{PDSketch}, which employs neural networks as the basic modules of human-specified programming structures and learns a transition model. This model supports generic network-based representations for predicates and action effects. Nevertheless, its model-based planning framework tends to accumulate errors, making it less suitable for long-horizon decision-making tasks.

In this work, we borrow the idea of abductive learning and introduce a novel framework \textbf{AB}ductive \textbf{I}mitation \textbf{L}earning (ABIL), which combines the benefits of data-driven learning and symbolic-based reasoning, enabling long-term planning based on state observations. 
Specifically, ABIL employ abductive reasoning to help understand the demonstrations in symbolic space and apply the principles of sequential consistency to resolve the conflicts between perception and reasoning. It applies logical reasoning to generate predicate candidates that meet constraints, eliminating the need for laborious symbolic annotations. 
With the above symbolic understanding, we further build a policy ensemble whose base policies are built with different logical objectives and managed by symbolic reasoning. 
The learned policy imitates specific behaviors directly from demonstrations, eliminating the reliance on prior low-level controllers used in earlier neuro-symbolic methods. Additionally, it makes decisions based on human-like cognition, which enhances its generalization capabilities.
  Experiments show that our proposal successfully understands the observations with the task-relevant symbolics to assist the imitating. 
  Notably, ABIL shows significantly improved performance in data efficiency and generalization settings across a variety of long-horizon tasks.

\begin{figure*}[t]
  \subfloat[Graphical knowledge of Cleaning a Car.]
  {\label{kb}\includegraphics[width=.98\columnwidth]{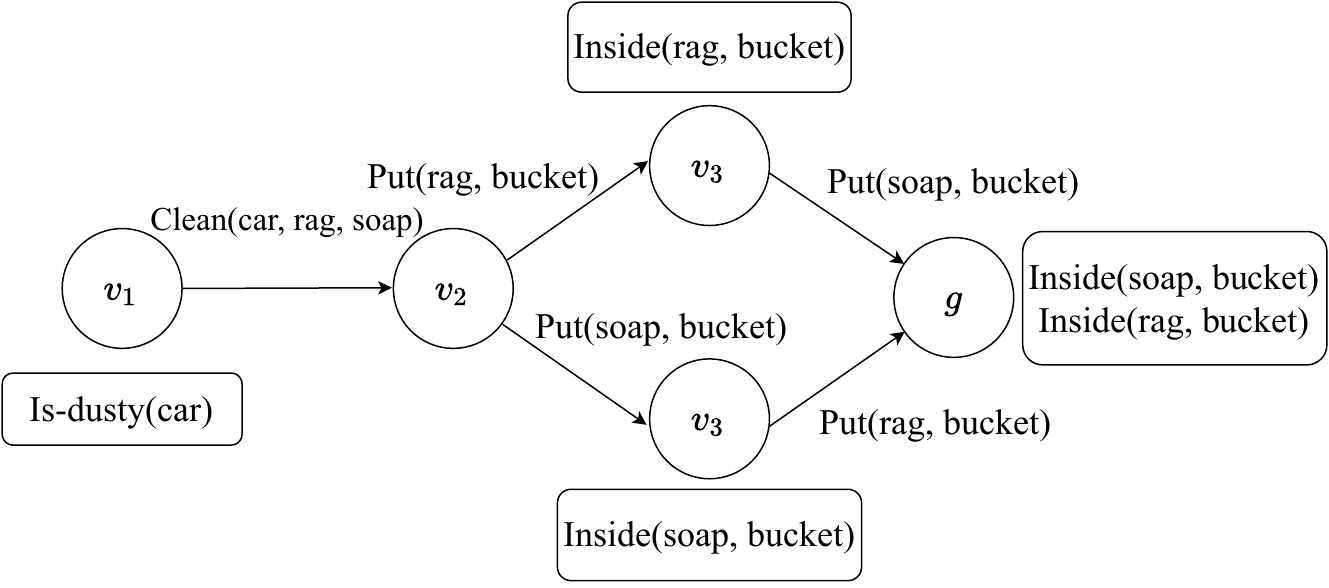}}
  \subfloat[The typical graphical structures of the state machine]{\label{fsm}
  \includegraphics[width=.98\columnwidth]{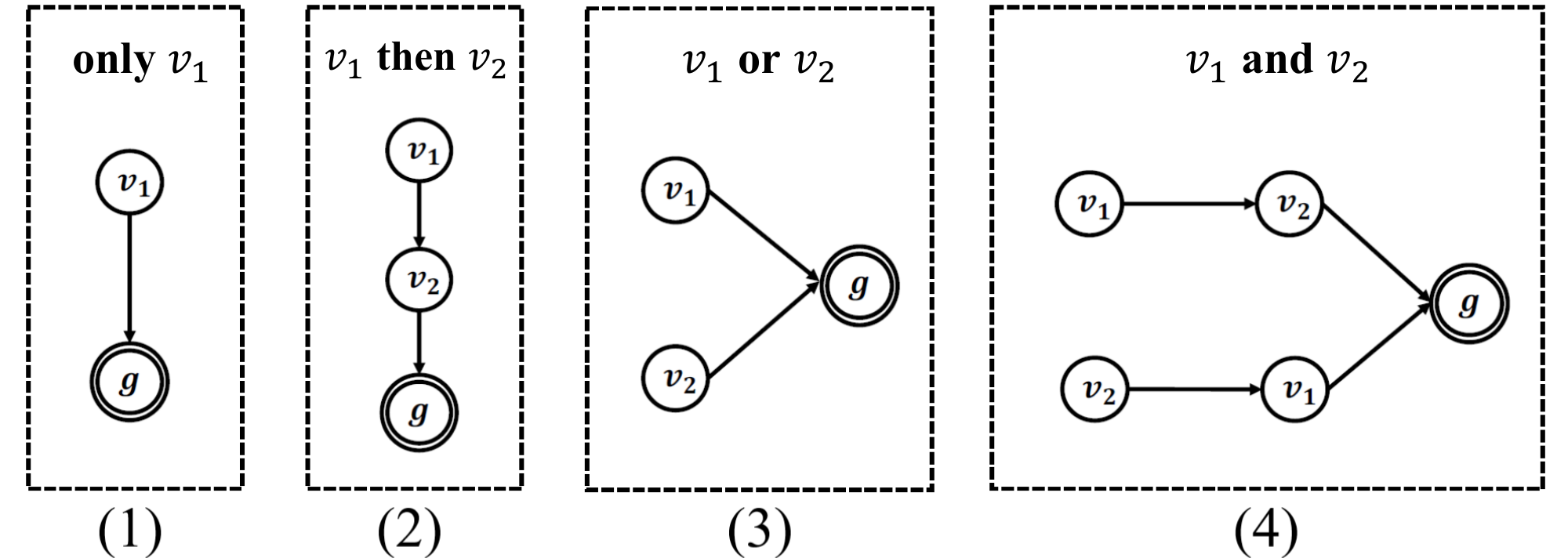}}
 \caption{Illustration of the Knowledge Base.}
\end{figure*}
\section{Related Work}
The preface work of this paper mainly includes \textit{Imitation Learning}, \textit{Neuro-Symbolic Planning} and \textit{Abductive Learning}. 

\textbf{Imitation Learning} 
learns the policies from expert demonstrations to achieve sequential decision-making~\cite{hussein2017imitation}. 
There are many works that obtain successful results on varying domains, such as robotic manipulation~\cite{DBLP:journals/ijira/FangJGXWS19, MimicPlay}, autonomous driving~\cite{DBLP:journals/tits/MeroYDM22, DBLP:journals/tits/BhattacharyyaWPKMSK23} and language models~\cite{whitehurst1975language, DBLP:conf/nips/BrownMRSKDNSSAA20}. 
However, learning theory reveals that the generalization ability of imitation learning is constrained by the size of the expert dataset and degrades as the decision-making horizon increases~\cite{DBLP:conf/nips/RajaramanYJR20,shao2024offline,DBLP:conf/kdd/ShaoSGL24}. 
This issue is particularly pronounced in open environments, where home agents need to accomplish tasks in differently arranged rooms. In such settings, the distribution shift, that is, the covariate shift between training observations and the scenarios the agent actually encounters, presents a greater challenge for imitation learning~\cite{Mini-BEHAVIOR, DBLP:conf/corl/0002ZWGSMWLLSAH22}. 

\textbf{Neuro-Symbolic Planning} 
explores to combine traditional symbolic planning with learning to enhance model's generalization capabilities. 
To handle observations beyond symbolic states, previous studies typically involve training neural networks with task-relevant predicate annotations to transform raw observations into symbolic states for planning. 
For example, \citeauthor{RPN}~\cite{RPN} propose the Regression Planning Networks, which learns to predict sub-goals that need to be achieved before the final goals, enabling traditional symbolic regression to handle complex high-dimensional inputs, like images. 
\citeauthor{DBLP:journals/jair/KonidarisKL18}~\cite{DBLP:journals/jair/KonidarisKL18} collect feasibility annotations and transition data under different symbolic operations to learn the symbolic representation of different observations. 
\citeauthor{DBLP:conf/iros/SilverCTKL21}~\cite{DBLP:conf/iros/SilverCTKL21} formalize operator learning for neuro-symbolic planning, viewing operators as an abstraction model of the raw transition and generating the high-level plan skeletons. 
\citeauthor{DBLP:conf/iclr/WangMHZ0G23}~\cite{DBLP:conf/iclr/WangMHZ0G23} leverage a pre-trained vision-language model to provide the predicate-level annotations 
to help imitation learning. 
The most related work to ours is PDSketch~\cite{PDSketch}, which utilizes neural networks as the basic modules of human-specified programming structures and learns an object-factored transition model that supports generic neural-network-based representations for predicates and action effects. 
However, we find that its planning, based on the raw-observation-action space, tends to accumulate errors and is not suitable for long-sequence decision-making tasks.
Furthermore, its application of logical reasoning is rather limited and usually requires large amounts of training data to achieve neuro-symbolic grounding. 
In contrast, we develop sequential consistency for abductive reasoning which results in a more data-efficient grounding, which assists imitation learning in turn. 

\textbf{Abductive Learning} provides a framework that integrates machine learning with logical reasoning~\cite{ABL_Zhou19, ABL_DaiX0Z19, HuangDCMJ21, HuangSLTD0JZ23, DBLP:conf/icml/YangWSLZ24}. It focuses on handling the intermediate neuro-symbolic grounding, which serves as pseudo-labels for learning and as variables for abduction. 
Although there have been some efforts to extend abductive learning to different applications, such as judicial sentencing~\cite{HuangDYCCHLZ20} and historical document understanding~\cite{ABL_hd}, they mainly consider the traditional classification tasks. 
In this work, we focus on the planning problems, where long-horizon sequential decision-making tasks are mainly considered. 

Generally speaking, it is still challenging to implement imitation in the real world using the above technologies. 
Considering that humans can relatively easily provide a knowledge base with high-level symbolic solutions for long-horizon decision-making tasks like robotic manipulation~\cite{DBLP:journals/jair/KonidarisKL18} or household tasks~\cite{RPN, Mini-BEHAVIOR, PDSketch}, this work follows the research line of neural-symbolic methods to use a knowledge base to assist imitation learning, while also avoiding the requirements on tedious predicate annotations. 

\begin{figure*}[t]
  \centering
  \includegraphics[width=1\linewidth]{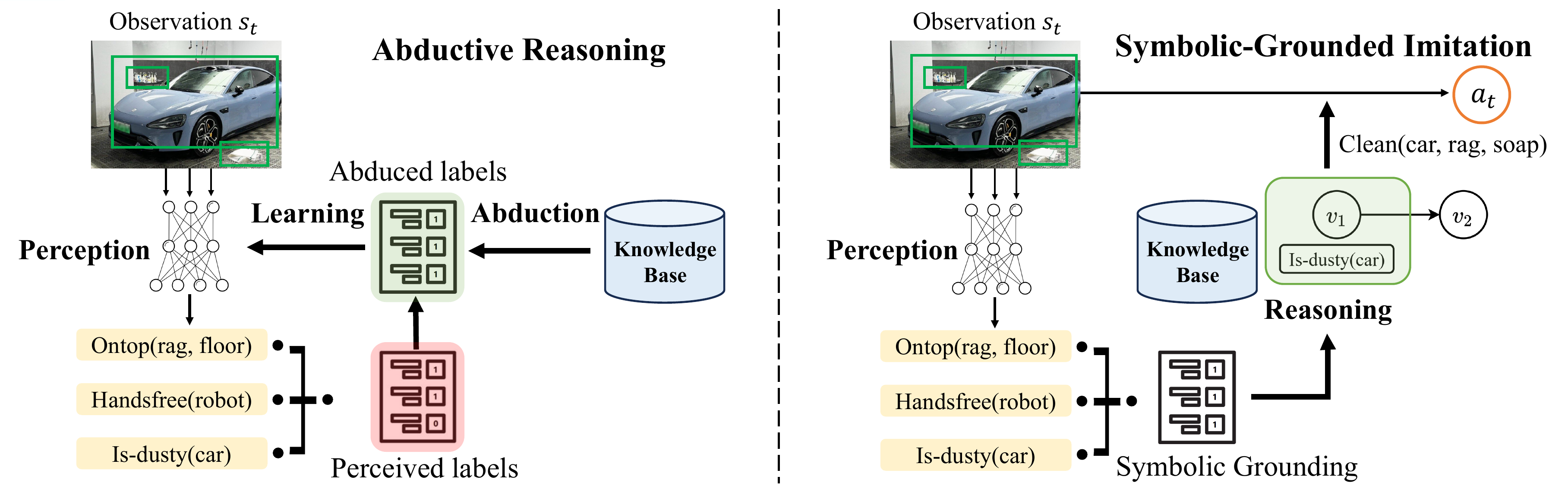}
  \caption{The Framework of our Abductive Imitation Learning.}
  \label{framework_fig}
\end{figure*}
\section{The Proposed Framework}

\subsection{Problem Formulation}
In this paper, we focus on the goal-based planning task. 
Following~\cite {PDSketch, DBLP:conf/corl/SilverATLK22, DBLP:conf/aaai/SilverCKMLKT23}, 
the environment is formally defined as a tuple $ \langle\mathcal{S},\mathcal{A},\mathcal{T}, \mathcal{O},\mathcal P,\mathcal{OP}, \mathcal{S}^0,g  \rangle$. Here, $\mathcal{S}$ represents the state space, and $\mathcal{A}$ represents the action space. The transitions between states and actions are governed by a deterministic transition function $\mathcal{T}$: $\mathcal{S}\times \mathcal{A}\to \mathcal{S}$. $\mathcal{S}^0$ denotes the distribution of initial states.
The set $\mathcal{O}$ consists of a finite number of task-related objects, where each object $o$ possesses a unique name, such as \textit{car} and \textit{rag}. Additionally, $\mathcal{P}$ is a finite set of task-related predicate symbols, where each predicate symbol $p$ has an arity that indicates the number of arguments it takes. For example, \textit{Inside/2} has an arity of two, representing that one object is inside another.
A ground atom $\overline p$ is a predicate that only contains concrete objects, such as \textit{Inside(rag, bucket)}. If a state $s$ satisfies $s\models \overline p$, it indicates that $s$ semantically entails the interpretation of $\overline p$.
The set $g$ consists of ground atoms that represent the task's target. The task is to find an action sequence that generates a trajectory $(s^0,a^1,s^1,...,a^T,s^T)$ satisfying $\forall \overline p \in g, s^T \models \overline p$. For simplification, we denote $s\models P$ (where $P$ is a set of ground atoms) as $\forall \overline p \in P, s \models \overline p$. Moreover, $\mathcal{OP}$ is also a set of finite predicate symbols that represent logical operators, such as \textit{clean/3} and \textit{put/2}. We denote $\textit{obj}(\overline p)$ to retrieve the object(s) from a ground atom. For instance, $\textit{obj}(\textit{Put(rag, bucket)})=\{\textit{rag, bucket}\}$. 

The symbolic knowledge base provided by experts could be formulated as a finite-state machine with a directed graph $G=\langle V, E \rangle$. Each node $v$ in the vertex set $V$ contains a set of ground atoms of $\mathcal{P}$, which can be viewed as the condition of a sub-task. Each edge is noted as a tuple $\left \langle \overline{op}, \text{EFF}^+,\text{EFF}^-\right \rangle$. $\overline{op}$ is a ground atom of $\mathcal{OP}$ representing the symbolic action, e.g., \textit{Put(rag, bucket)}. $\text{EFF}^+$ is the add effect and $\text{EFF}^-$ is the delete effect, each is a set of grounding atoms. 
A symbolic action $op$ typically requires multiple actions $a\sim \mathcal{A}$ to achieve the desired logical sub-goal, corresponding to a segment of the complete trajectory. 
For the sake of simplicity, the notation $\overline{\mathrm{op}}$ is utilized to denote the edge. 

For each node $u,v\in V$, if there is a directed edge pointing from $u$ to $v$, then $v=(u-\text{EFF}^-)\cup \text{EFF}^+$. We define a trajectory $z=(s^0,a^1,s^1,...,a^T,s^T)$ satisfying the knowledge base $G$ denoted as $z\models G$ if and only if for every adjacent states pair $(s^t,s^{t+1})$, there exists $u\in V, s^t\models u \wedge s^{t+1} \models u$ or $s^t\models u$, $s^{t+1}\models v, \exists u,v\in V, (u,v)\in E$. 
This indicates that the expert trajectory satisfies the corresponding symbolic knowledge base, such as first using a rag and soap to clean a dusty car, and then putting them into a bucket.
Figure \ref{kb} demonstrates an example of the knowledge base formalized as a state machine with a directed graph.

The state machine contains multiple basic structures as illustrated in Figure \ref{fsm}. In the event that a singular node, denoted as $v_1$, directs towards the goal, it signifies the necessity to address a corresponding sub-task $v_1$. For instance, the action \textit{Clean(car, rag, soap)} is imperative whenever the objective is to clean a car. Conversely, should there be a directed edge from $v_1$ to $v_2$, with $v_2$ subsequently pointing towards the goal, it implies that the sub-task $v_1$ must be solved before sub-task $v_2$. In scenarios where both $v_1$ and $v_2$ have directed edges towards the goal, either sub-task $v_1$ or $v_2$ is required to be solved. The final configuration arises when there is bidirectional pointing between $v_1$ and $v_2$, indicating that both sub-tasks are mandatory to be solved. Utilizing the state machine, a symbolic planning solution can be derived via algorithms such as $A^*$ search or dynamic programming, represented as a sequence of states and transitions: $\{(v_0, \overline{op}_0), (v_1, \overline{op}_1)\dots (v_K, \overline{op}_K)\}$.

\subsection{Abductive Reasoning}

The ABIL framework can be roughly divided into abductive learning with the state machine and imitation with symbolic reasoning. 
The overall framework is illustrated and summarized in Figure~\ref{framework_fig} and Algorithm~\ref{algo}, respectively. 

Given the state machine and expert demonstrations, the challenge lies in establishing the perception function $f$ from observation to symbolic grounding when symbolic supervision of $\{s_t\}$ is not available. To address this challenge, we introduce abductive reasoning to provide pseudo labels derived from the state machine's knowledge, which could be taken to optimize the perception function $f$.

\begin{equation}
  \begin{aligned}
  \min_f &\sum_{s_i \in D} \sum_{t=1}^T \mathcal{L}(f(s_i^t), \widehat{z_i^t}), \\
  \{\widehat{z_i^t}\}_{t=1}^T =\arg\min_{\{z_i^t\}_{t=1}^T} & \sum \lVert z_i^t-f(s_i^t)\rVert^2, \quad \text{s.t.} \{z_i^t\}_{t=1}^T \models G
  \label{abl_eq}
  \end{aligned}
\end{equation}

Specifically, for the different structures in the state machine, we could derive the sequential abduction: $\{z_i^t\}_{t=1}^T \models G$ as: 
\begin{itemize}
  \item The task has been completed at the end of the expert demonstration sequence. The symbolic state $z_i^T$ satisfies the final goal $g$, that is, $z_i^T\models g$. 
  \item Task $a$ should be accomplished before $b$. The symbolic sequence $\{z_i^t\}_{t=1}^T$ will satisfy $\{z_i^t\}_{t=1}^{j}\models a, \{z_i^t\}_{t=j+1}^{T}\models b, \exists j\in (1,T)$ where the demonstrations could be divided into the two sequential part of accomplishing $a$ and $b$. 
  \item The agent should either complete $a$ or $b$. The symbolic sequence $\{z_i^t\}_{t=1}^T$ will satisfy $\{z_i^t\}_{t=1}^{T}\models a$ or $\{z_i^t\}_{t=1}^{T}\models b$.  
  \item The agent should complete both t1 and t2, but in any order. The symbolic sequence $\{z_i^t\}_{t=1}^T$ will satisfy $\{z_i^t\}_{t=1}^{j}\models a, \{z_i^t\}_{t=j+1}^{T}\models b, \exists j\in (1,T)$ or $\{z_i^t\}_{t=1}^{j}\models b, \{z_i^t\}_{t=j+1}^{T}\models a, \exists j\in (1,T)$.
\end{itemize}

\begin{algorithm}[tb]
  \caption{Abductive Imitation Learning}
  \label{algo}
  \begin{algorithmic}[1]
  \REQUIRE Demonstration dataset $D$, symbolic knowledge $G$. Number of learning rounds $N_R$ and $N_I$. 
  \FOR{$t=1$ to $N_R$}
    \STATE Get the perceived labels via $f(s)$
    \STATE Get the abduced labels via Eq.~\ref{abl_eq}. 
    \STATE Update the perception network $f$. 
  \ENDFOR
  \FOR{$t=1$ to $N_I$}
    \STATE Get the symbolic states via $f(s)$
    \STATE Get the logical operator $\bar{op}$ via Eq.~\ref{get_op}. 
    \STATE Update the behavior network $h_{\bar{op}}$ via Eq.~\ref{update_actor}. 
  \ENDFOR
  \STATE \textbf{return} Perception $f$ and behavior $\{h_{\bar{op}}\}, \bar{op}\in \mathcal{OP}$.
  \end{algorithmic}
\end{algorithm}
With the sequential abduction, the pseudo label $\hat{z}$ in Equation~\ref{abl_eq} could be obtained and the perception module $f$ could be optimized. Nevertheless, the perception module $f$ plays an important role in symbolic grounding, 
which is crucial for the subsequent reasoning. 
Following~\cite{DBLP:conf/iros/HuangXZGSFN19,PDSketch, DBLP:conf/corl/SilverATLK22, DBLP:conf/aaai/SilverCKMLKT23}, we train the perception module $f$ at the object level. For each predicate $p$, we have a predicate model $f_p$ with the object-level features $o$ as input, which could be obtained via an object detection model in practice. A ground atom $p(o1,o2)$ could be inferred by: $\textit{prob}(p(o_1, o_2)|s)= f_p ([o_1, o_2])$. 

As summarized in the left part of Figure~\ref{framework_fig}, our perception module $f$ is optimized via the abductive reasoning. Equation~\ref{abl_eq} generates the pseudo labels based on the sequential consistency between the perception output and the solution of $z$ satisfying the knowledge base. Unlike previous works~\cite{RPN, DBLP:conf/iclr/WangMHZ0G23}, it does not rely on the symbolic-level annotations of each observation in demonstrations, which are usually costly and difficult to obtain.

\subsection{Symbolic-grounded Imitation} 
As a human being, one often consciously knows what he or she is doing, such as making the action of turning left because they realize the destination is on the left. In this part, we thus incorporate the reasoning of high-level operators into the original imitation learning process, regarding the symbolic operators as an assistance signals. 
Specifically, we first build the behavioral actor for each logical operator $h_{op}$, e.g. $h_{clean}$ and $h_{put}$. Then we derive the desired behavior module by the symbolic states output of perception $f$ and the corresponding abstract logical operator. Given the solution of symbolic planning: $\{(v_0, \overline{op}_0), (v_1, \overline{op}_1)\dots (v_K, \overline{op}_K)\}$. 
\begin{equation}
  \overline{op}^t = \overline{op}_k, \text{s.t.} f(s^t) \models v_k, \exists k \in [0,K)
  \label{get_op}
\end{equation}
The desired parameter of the operator $\overline{op}^t $, e.g., which object will be picked, could be reasoning as: 
\begin{equation}
  o^t = obj(\overline{op}^t) 
\end{equation}
Then the learning of behavior actors could be formulated as: 
\begin{equation}
  \min_h \sum_{s_i,a_i\in D} \sum_{t=1}^T \mathcal{L}(h_{\overline{op}_i^t}(s_i^t, o^t), a_i^t)
  \label{update_actor}
\end{equation}
As summarized in the right part of Figure~\ref{framework_fig}, our behavioral actors, referring to the human model of cognition before decision-making, embed high-level logical reasoning into the imitation learning process. 
The behavior ensemble $h_{\bar{op}}$ learns from experience through imitation, without relying on a pre-existing perfect controller to reach each sub-goal. Importantly, by leveraging the generalization capabilities of symbolic planning, the proposed actors can decompose diverse observations into symbolic states, facilitating more reliable decision-making. 

\begin{table*}[t]
  \caption{Success rates@100 in the BabyAI benchmark. The best result in each setting is bold and the second is underlined.}
  \vspace{-0.1in}
  \setlength{\tabcolsep}{3mm}{\begin{tabular}{lc|ccccc}
  \toprule
  Task \hfill(Averaged Length) &Eval& BC & DT &PDSketch& ABIL-BC & ABIL-DT  \\
  \midrule
  GotoSingle \hfill(3)  &Basic & \textbf{1.00} & 0.893$\ \pm\ $0.049&\textbf{1.00} & \textbf{1.00}&\underline{0.963$\ \pm\ $0.047}\\
  \midrule
  \multirow{2}*{Goto }\hfill\multirow{2}*{(3)} &Basic& 0.843$\ \pm\ $0.006& 0.720$\ \pm\ $0.044 &\textbf{1.00} &0.900$\ \pm\ $0.046& \underline{0.900$\ \pm\ $0.020} \\
  &Gen& 0.743$\ \pm\ $0.045&0.583$\ \pm\ $0.049&\textbf{1.00} & 0.777$\ \pm\ $0.032& \underline{0.793$\ \pm\ $0.029}\\
  \midrule
  \multirow{2}*{Pickup }\hfill\multirow{2}*{(4)}  &Basic   & 0.723$\ \pm\ $0.031& 0.490$\ \pm\ $0.040 &\textbf{0.990$\ \pm\ $0.010} & \underline{0.847$\ \pm\ $0.025} & 0.845$\ \pm\ $0.035\\
  &Gen & 0.533$\ \pm\ $0.031& 0.320$\ \pm\ $0.070 &\textbf{0.973$\ \pm\ $0.012}& 0.730$\ \pm\ $0.010& \underline{0.763$\ \pm\ $0.051}\\
   \midrule
  \multirow{2}*{Open }\hfill\multirow{2}*{(6)}  &Basic  & 0.933$\ \pm\ $0.025&   0.493$\ \pm\ $0.059 &\textbf{1.00}   & \underline{0.963$\ \pm\ $0.021}  &0.923$\ \pm\ $0.031 \\
  &Gen& 0.877$\ \pm\ $0.015&   0.440$\ \pm\ $0.078&\textbf{1.00} & \underline{0.927$\ \pm\ $0.032} & 0.857$\ \pm\ $0.055\\
  \midrule
  \multirow{2}*{Put }\hfill\multirow{2}*{(9)}    &Basic  &\underline{0.950$\ \pm\ $0.044}& 0.910$\ \pm\ $0.036&0.697$\ \pm\ $0.021& 0.940$\ \pm\ $0.026 & \textbf{0.953$\ \pm\ $0.006}\\
   &Gen & 0.260$\ \pm\ $0.036& 0.380$\ \pm\ $0.026& 0.417$\ \pm\ $0.025& \textbf{0.637$\ \pm\ $0.064}& \underline{0.593$\ \pm\ $0.084} \\
    \midrule
  \multirow{2}*{Unlock }\hfill\multirow{2}*{(10)}   &Basic    & 0.957$\ \pm\ $0.012& \underline{0.990$\ \pm\ $0.010}&0.293$\ \pm\ $0.051& 0.967$\ \pm\ $0.023 & \textbf{0.993$\ \pm\ $0.012}\\
   &Gen & 0.910$\ \pm\ $0.030 & \underline{0.990$\ \pm\ $0.010} & 0.247$\ \pm\ $0.051& 0.963$\ \pm\ $0.006 & \textbf{0.993$\ \pm\ $0.012}\\
   \midrule
  \multicolumn{2}{l|}{Averaged time per evaluation}&0.174 seconds & 0.260 seconds&8.170 seconds&0.320 seconds&0.354 seconds\\
  \bottomrule
  \end{tabular}\label{tab:babyai}}
  \end{table*}

\begin{figure*}[th]
  \vspace{-0.15in}
  \centering
  \subfloat[Goto]
  {
\label{grounding:subfig1}\includegraphics[width=0.24\linewidth]{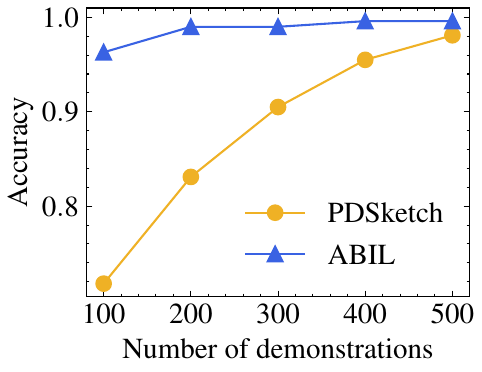}
  }
  \subfloat[Pickup]
  {
\label{grounding:subfig2}\includegraphics[width=0.24\linewidth]{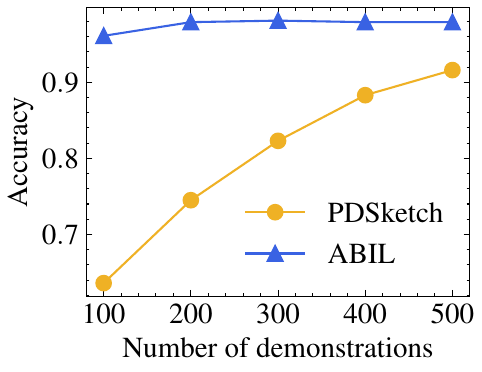}
  }
   \subfloat[Open]
  {
\label{grounding:subfig3}\includegraphics[width=0.24\linewidth]{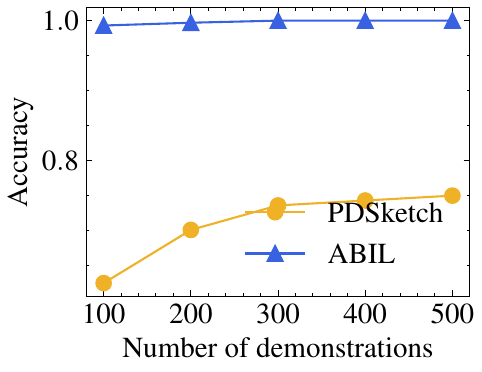}
  }
  \subfloat[Unlock]
  {
\label{grounding:subfig4}\includegraphics[width=0.24\linewidth]{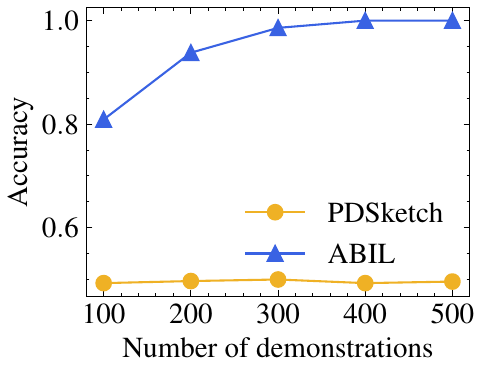}
  }
  \vspace{-0.1in}
  \caption{Accuracy of neuro-symbolic grounding under varying data budgets.}
  \vspace{-0.1in}
  \label{grounding_efficiency}
\end{figure*}

\section{Empirical Study}
We evaluate our proposal in three environments, including two neuro-symbolic benchmarks: BabyAI~\cite{BabyAI-Chevalier-Boisvert19}, Mini-BEHAVIOR~\cite{Mini-BEHAVIOR}, and a robotic manipulation benchmark, CLIPort~\cite{cliport}. We compare our method with three baselines: Behavior Cloning (\textbf{BC})~\cite{BC}, Decision Transformer (\textbf{DT})~\cite{DT2021} and \textbf{PDSketch}~\cite{PDSketch}. 
For a fair comparison, all of these methods use the same network architecture which is based on the Neural Logic Machine~(NLM)~\cite{NLM}. 
Specifically, we first encode the state with a two-layer NLM. For BC, we use a single linear layer, taking the state embedding as the input and output actions. For DT, we build a single transformer layer following the two-layer encoder, with the causal mask to generate future action with past states and actions. 
For PDSketch, we choose the full mode in the original paper~\cite{PDSketch}, which provides sufficient prior knowledge of the symbolic transition, keeping consistency with our symbolic state machine. Following~\cite{PDSketch, cliport}, we report the percentage of successful planning for the desired goals, which are averaged over 100 evaluations under three random seeds.

\subsection{Evaluation on BabyAI}
BabyAI provides a benchmark 
for grounding logical instructions 
where an agent needs to follow 
instructions for a series of tasks like picking up objects, or unlocking doors.  
Following~\cite{RPN, PDSketch}, we consider 5 tasks: $\{goto, pickup, open, put, unlock\}$, and conduct the generalization evaluation with different numbers of objects in the testing environments. For example, the training environments contain 4 objects or 4 doors, and testing environments contain 8 objects or 8 doors. It simulates the challenges to the generalization of imitation learning, where household robots in open environments need to complete tasks in differently arranged rooms. All demonstrations of the expert dataset are generated by a script based on $\mathit{A}^* $ search.

\textbf{Results and Analysis.}
From Table~\ref{tab:babyai}, we could find that
BC baseline is efficient in the simple \textit{GotoSingle} task but significantly deteriorates on complex tasks and their generalization evaluation. 
PDSketch exhibits favorable performance in tasks that require few actions to complete and generalizes well in the case of increasing object number. Nevertheless, it struggles to solve long-horizon tasks like \textit{Put} and \textit{Unlock}, where experts’ demonstrations require 9 or 10 actions to accomplish. 
One plausible reason is that, as a model-based method, PDSketch faces the accumulation of search errors, thus fails in long-horizon tasks. 
DT excels in handling tasks with a sequential nature (e.g. \textit{Put} and \textit{Unlock}). 
However, in short-horizon tasks, such as \textit{Goto} and \textit{Pickup}, DT performs weaker compared to BC. The possible reason could be its relatively high model complexity contradicts with limited data, making it difficult to learn efficiently from short-horizon demonstrations. 
In contrast, ABIL achieves competitive performance compared to PDSketch and has made significant improvements in long-horizon tasks. 
It is worth noting that ABIL exhibits stable generalization in novel environments, 
which supports our intention of using symbolic grounding to assist the generalization of imitation learning. 

\begin{table*}[t]
\caption{Success rates@100 in the Mini-BEHAVIOR benchmark.  The best result in each setting is bold. }
\setlength{\tabcolsep}{2.7mm}{\begin{tabular}{lc|ccccc}
\toprule
Task \hfill (Averaged Length) &Eval& BC  & DT & PDSketch& ABIL-BC&ABIL-DT\\
\midrule
\multirow{2}*{Installing a printer}\hfill\multirow{2}*{(10)}  &Basic   & 0.903$\ \pm\ $0.023& 0.927$\ \pm\ $0.021& 0.343$\ \pm\ $0.032 & 0.920$\ \pm\ $0.035&\textbf{0.947$\ \pm\ $0.012}\\
&Gen& 0.003$\ \pm\ $0.006& 0.300$\ \pm\ $0.147 & 0.310$\ \pm\ $0.046 & 0.577$\ \pm\ $0.167&\textbf{0.760$\ \pm\ $0.010}\\
\midrule
\multirow{2}*{Opening packages}\hfill\multirow{2}*{(19)}  &Basic    & 0.947$\ \pm\ $0.045& 0.963$\ \pm\ $0.034&0.020$\ \pm\ $0.010 &0.988$\ \pm\ $0.016 & \textbf{0.993$\ \pm\ $0.008}\\
&Gen&0.295$\ \pm\ $0.180 & 0.548$\ \pm\ $0.065 &0.020$\ \pm\ $0.010& \textbf{0.892$\ \pm\ $0.043}& 0.845$\ \pm\ $0.071\\
\midrule
\multirow{2}*{Making tea}\hfill\multirow{2}*{(36)}   &Basic   &0.607$\ \pm\ $0.015 & 0.583$\ \pm\ $0.105 & \multirow{2}*{> 5 minutes}& 0.613$\ \pm\ $0.032 &\textbf{0.623$\ \pm\ $0.012}\\
&Gen& 0.070$\ \pm\ $0.078 & 0.113$\ \pm\ $0.105 & & 0.074$\ \pm\ $0.053 &\textbf{0.487$\ \pm\ $0.049}\\
\midrule
\multirow{2}*{Moving boxes to storage}\hfill\multirow{2}*{(38)}  &Basic    &0.783$\ \pm\ $0.061 & 0.787$\ \pm\ $0.060 & \multirow{2}*{> 5 minutes}& 0.803$\ \pm\ $0.031&\textbf{0.807$\ \pm\ $0.038}\\
&Gen& 0.433$\ \pm\ $0.470 & 0.613$\ \pm\ $0.047 & &0.717$\ \pm\ $0.049&\textbf{0.753$\ \pm\ $0.038}\\
\midrule
\multirow{2}*{Cleaning A Car }\hfill\multirow{2}*{(45)}&Basic & 0.417$\ \pm\ $0.047& 0.313$\ \pm\ $0.091&\multirow{2}*{> 5 minutes} & \textbf{0.420$\ \pm\ $0.036}&0.340$\ \pm\ $0.090 \\
&Gen& 0.170$\ \pm\ $0.036 & 0.147$\ \pm\ $0.083 & &0.183$\ \pm\ $0.104 &\textbf{0.220$\ \pm\ $0.017}\\
\midrule
\multirow{2}*{Throwing away leftovers}\hfill\multirow{2}*{(46)}  &Basic   &0.833$\ \pm\ $0.080 & 0.890$\ \pm\ $0.029 & \multirow{2}*{> 5 minutes} & \textbf{0.941$\ \pm\ $0.030}&0.855$\ \pm\ $0.059\\
&Gen& 0.222$\ \pm\ $0.167 & 0.653$\ \pm\ $0.039 & & 0.488$\ \pm\ $0.147&\textbf{0.680$\ \pm\ $0.039}\\
\midrule
\multirow{2}*{Putting away dishes}\hfill\multirow{2}*{(65)} &Basic     &0.811$\ \pm\ $0.031 & 0.828$\ \pm\ $0.052 &\multirow{2}*{> 5 minutes} & \textbf{0.872$\ \pm\ $0.024}&0.811$\ \pm\ $0.018\\
&Gen& 0.141$\ \pm\ $0.111 & 0.547$\ \pm\ $0.296 && \textbf{0.758$\ \pm\ $0.118}&0.679$\ \pm\ $0.108\\
\midrule
\multirow{2}*{Sorting books}\hfill\multirow{2}*{(66)} &Basic     & 0.601$\ \pm\ $0.032 & 0.543$\ \pm\ $0.053 & \multirow{2}*{> 5 minutes}& \textbf{0.672$\ \pm\ $0.084}&0.575$\ \pm\ $0.011\\
&Gen& 0.131$\ \pm\ $0.047 & 0.220$\ \pm\ $0.010&  & \textbf{0.360$\ \pm\ $0.037}&0.333$\ \pm\ $0.048\\
\midrule
\multirow{2}*{Laying wood floors}\hfill\multirow{2}*{(68)} &Basic    & 0.616$\ \pm\ $0.062& 0.638$\ \pm\ $0.027 & \multirow{2}*{> 5 minutes} & 0.641$\ \pm\ $0.054&\textbf{0.643$\ \pm\ $0.036}\\
&Gen& 0.068$\ \pm\ $0.018 & 0.366$\ \pm\ $0.041 &  & 0.225$\ \pm\ $0.134&\textbf{0.435$\ \pm\ $0.067}\\
\midrule
\multirow{2}*{Watering houseplants}\hfill\multirow{2}*{(68)} &Basic  &0.814$\ \pm\ $0.034 & 0.806$\ \pm\ $0.020 & \multirow{2}*{> 5 minutes}& \textbf{0.824$\ \pm\ $0.023}&0.812$\ \pm\ $0.034\\
&Gen&0.002$\ \pm\ $0.004 & 0.187$\ \pm\ $0.113 & & 0.197$\ \pm\ $0.095&\textbf{0.409$\ \pm\ $0.151}\\
\midrule
\multirow{2}*{Cleaning shoes}\hfill\multirow{2}*{(78)} &Basic   & 0.482$\ \pm\ $0.086& 0.427$\ \pm\ $0.042 & \multirow{2}*{> 5 minutes}& \textbf{0.623$\ \pm\ $0.033}&0.505$\ \pm\ $0.075\\
&Gen&0.030$\ \pm\ $0.005 & 0.053$\ \pm\ $0.046 &  & \textbf{0.215$\ \pm\ $0.106}&0.193$\ \pm\ $0.120\\
\midrule
\multirow{2}*{Collect misplaced items}\hfill\multirow{2}*{(86)} &Basic  & 0.460$\ \pm\ $0.030&   0.299$\ \pm\ $0.015  &\multirow{2}*{> 5 minutes}& \textbf{0.577$\ \pm\ $0.053} &0.421$\ \pm\ $0.042\\
&Gen& \textbf{0.325$\ \pm\ $0.074} & 0.261$\ \pm\ $0.023 & & 0.270$\ \pm\ $0.062 &0.279$\ \pm\ $0.020\\
\midrule
\multirow{2}*{Organizing file cabinet}\hfill\multirow{2}*{(106)}   &Basic   &0.156$\ \pm\ $0.047 & 0.522$\ \pm\ $0.067 & \multirow{2}*{> 5 minutes}& 0.166$\ \pm\ $0.028&\textbf{0.596$\ \pm\ $0.058}\\
&Gen&0.083$\ \pm\ $0.012 & 0.382$\ \pm\ $0.112 & & 0.109$\ \pm\ $0.032&\textbf{0.490$\ \pm\ $0.078}\\
\midrule
  \multicolumn{2}{l|}{Averaged time per evaluation}&1.48 seconds& 2.09 seconds&> 5 minutes &2.88 seconds&2.98 seconds\\
\bottomrule
\end{tabular}
\label{result_mini}}
\end{table*}
\begin{table*}[t]
\caption{Success rates@100 in the CLIPort benchmark. The best result in each setting is bold.}
\vspace{-0.05in}
  \setlength{\tabcolsep}{1mm}{\begin{tabular}{l ccccccccc}
  \toprule
  Task & Packing-5shapes & Packing-20shapes &  Placing-red-in-green & Putting-blocks-in-bowls &Separating-20piles&Assembling-kits  \\
\midrule
BC    &	0.883$\ \pm\ $0.025 & 0.207$\ \pm\ $0.006	& 0.840$\ \pm\ $0.031&0.507$\ \pm\ $0.030 &0.226$\ \pm\ $0.017 & 0.187$\ \pm\ $0.015\\
\midrule
DT   &0.913$\ \pm\ $0.046  & 0.180$\ \pm\ $0.026 &  0.846$\ \pm\ $0.024 & 0.539$\ \pm\ $0.068 & 0.250$\ \pm\ $0.048 & 0.177$\ \pm\ $0.023\\
\midrule
 ABIL-BC & \textbf{0.983$\ \pm\ $0.015} &\textbf{0.940$\ \pm\ $0.030} & 0.988$\ \pm\ $0.014&\textbf{0.962$\ \pm\ $0.012}& 0.305$\ \pm\ $0.011&\textbf{ 0.829$\ \pm\ $0.008}\\
\midrule
ABIL-DT   &0.977$\ \pm\ $0.021  & 0.857$\ \pm\ $0.025 & \textbf{0.989$\ \pm\ $0.017}&0.917$\ \pm\ $0.033&\textbf{0.382$\ \pm\ $0.029}&0.809$\ \pm\ $0.008\\
  \bottomrule
  \end{tabular}\label{tab:cliport}}
\vspace{-0.05in}
\end{table*}

\textbf{Comparison of Neuro-Symbolic Grounding.} 
Like previous neuro-symbolic methods, the efficiency of neuro-symbolic grounding is the key to determining whether successful or not.  
In Figure~\ref{grounding_efficiency}, we compare the accuracy of the predicates learned by ABIL and PDSketch, under varying demonstration budgets.
We found that in the relatively simple task, PDSketch can achieve over 90\% predicate accuracy with 500 demonstrations. However, in more challenging tasks, such as open and unlock, its neuro-symbolic grounding ability is quite poor, which also leads to unsatisfactory performance in Table~\ref{tab:babyai}. In contrast, our ABIL not only achieves reliable neuro-symbolic grounding on all tasks (with nearly perfect predicate accuracy), but it's also more data efficient than PDSketch, requiring less than 20\% of their data to achieve superior neuro-symbolic grounding results. It clearly indicates the advantages of our abductive reasoning on neuro-symbolic understanding. 

\textbf{Comparison of Efficiency.} 
Learning-based methods BC and DT could promptly provide responses in the inference phase because they do not give adequate attention to the subsequent considerations. The model-based planning method PDSketch needs to search for a whole sequence of actions that can achieve the goal, which can be time-consuming, especially in cases with longer sequences and a multitude of available actions. 
Our approach integrates higher-level reasoning into the foundation of lower-level perception. 
By integrating symbolic-based planning, our approach enhances planning effectiveness, significantly reducing time consumption compared to PDSketch, which requires searching in the original observation-action space. 
While gaining advantages of logical reasoning in long-horizon goals, ABIL maintains the inference efficiency of the learning-based method. 

\subsection{Evaluation on Mini-BEHAVIOR}
Mini-BEHAVIOR is a recently proposed benchmark for embodied AI. It contains 
varying 3D household tasks chosen from the BEHAVIOR benchmark, including \textit{Sorting Books}, \textit{Making Tea}, \textit{Cleaning A Car}, and so on. Most tasks are long-horizon and heterogeneous, some of which require more than one hundred decision-making steps to be completed. 
There are hundreds of different types and plenty of predicates which is challenging for neuro-symbolic grounding. 
In this domain, our state machine is mainly composed of several typical categories. For tasks mainly about tidying up the room, we split the primitive actions into $op_{pick}$ and $op_{place}$, which is required to perform an action sequence to finish picking or placing sub-tasks. Combined with our symbolic-grounding model $f$, the agent will be able to distinguish when and where to pick and place. 
For tasks mainly about cleaning, we split the primitive actions into $op_{clean}$ and $op_{put}$, which is required to finish washing or putting sub-tasks. 
In the generalization evaluation, we challenge the agents in environments with distractor objects that are unseen at the training phase. 
\begin{figure*}[t]
  \centering
  \begin{minipage}[b]{2\columnwidth}
    \centering
    {\includegraphics[width=0.75\linewidth]{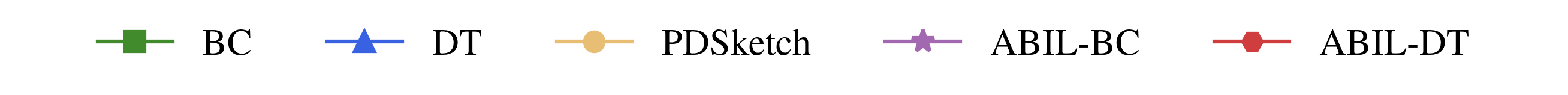}}
    \vspace {-0.175in}
\end{minipage}
\vspace {-0.05in}
   \subfloat[Pickup (Basic/Gen)]
  {
\label{fig:subfig1}\includegraphics[width=0.34\linewidth]{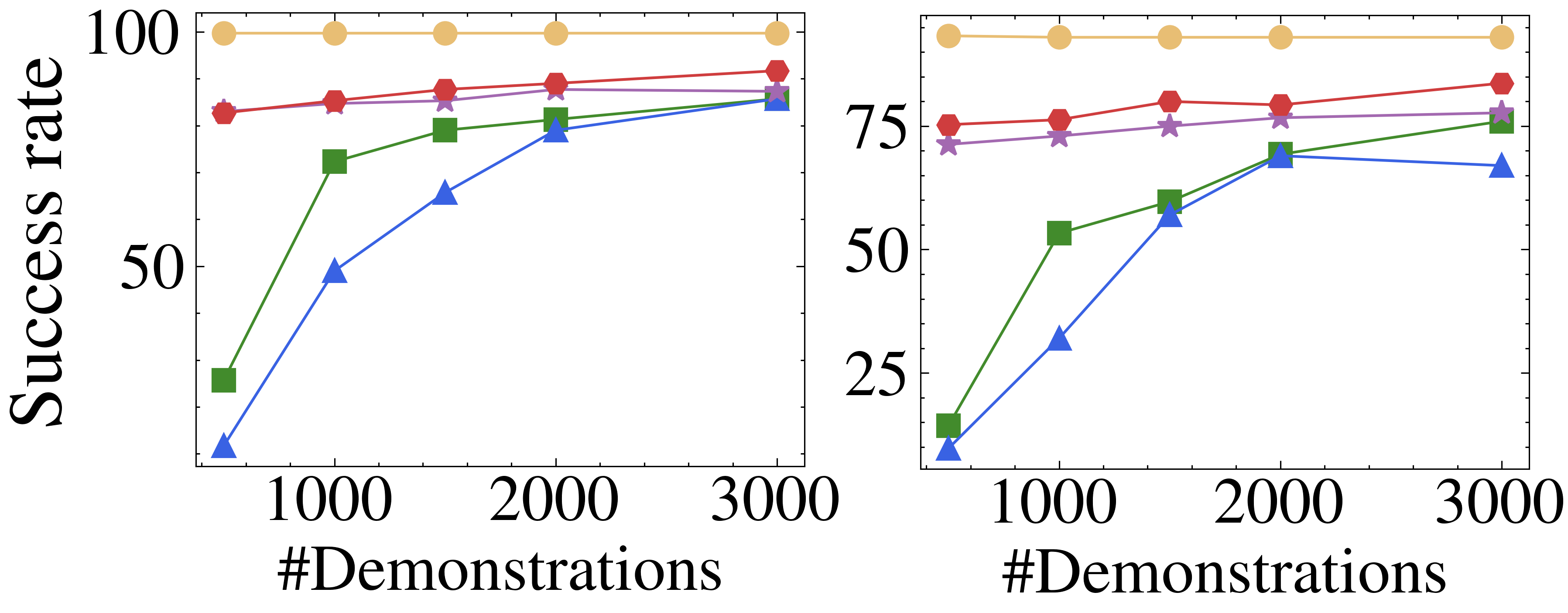}
  }
  \subfloat[Opening Packages(Basic/Gen)]
  {
\label{fig:subfig2}\includegraphics[width=0.325\linewidth]{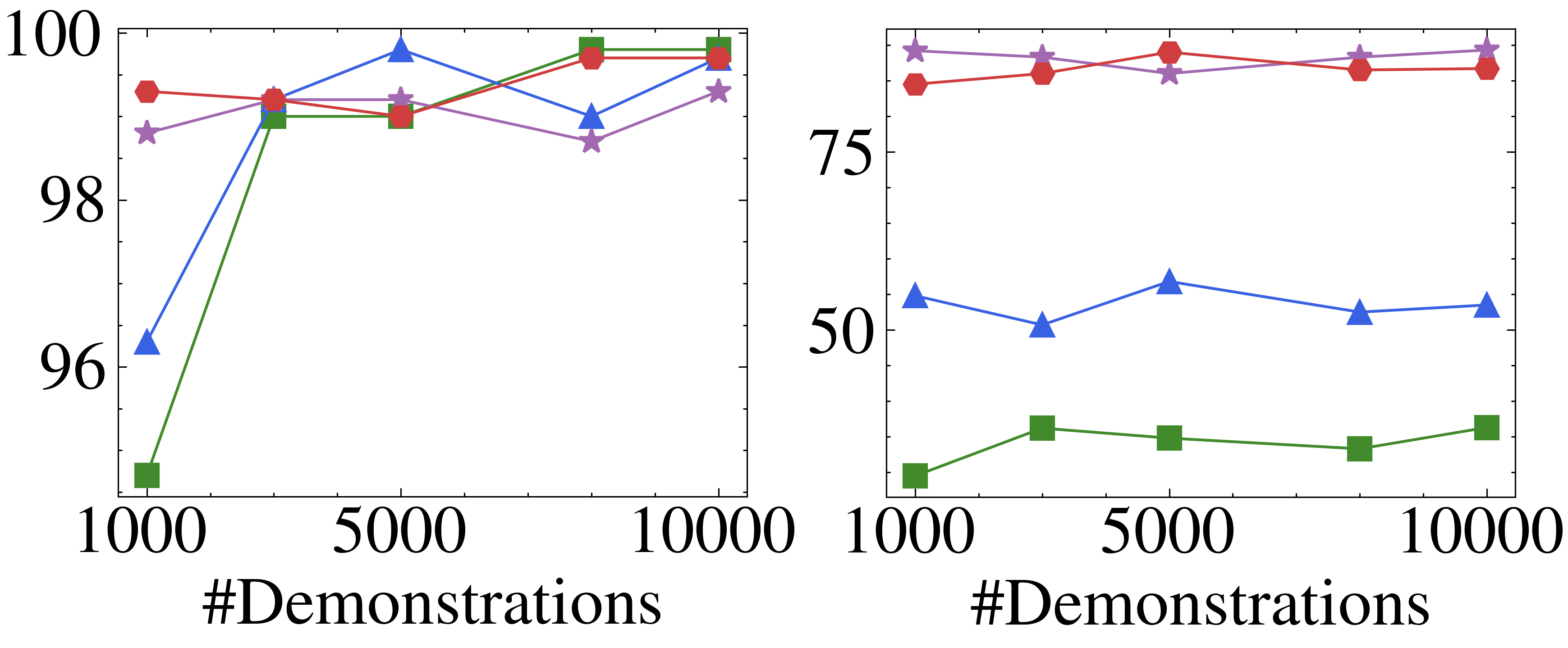}}
\subfloat[Putting dishes(Basic/Gen)]
  {
\label{fig:subfig3}\includegraphics[width=0.32\linewidth]{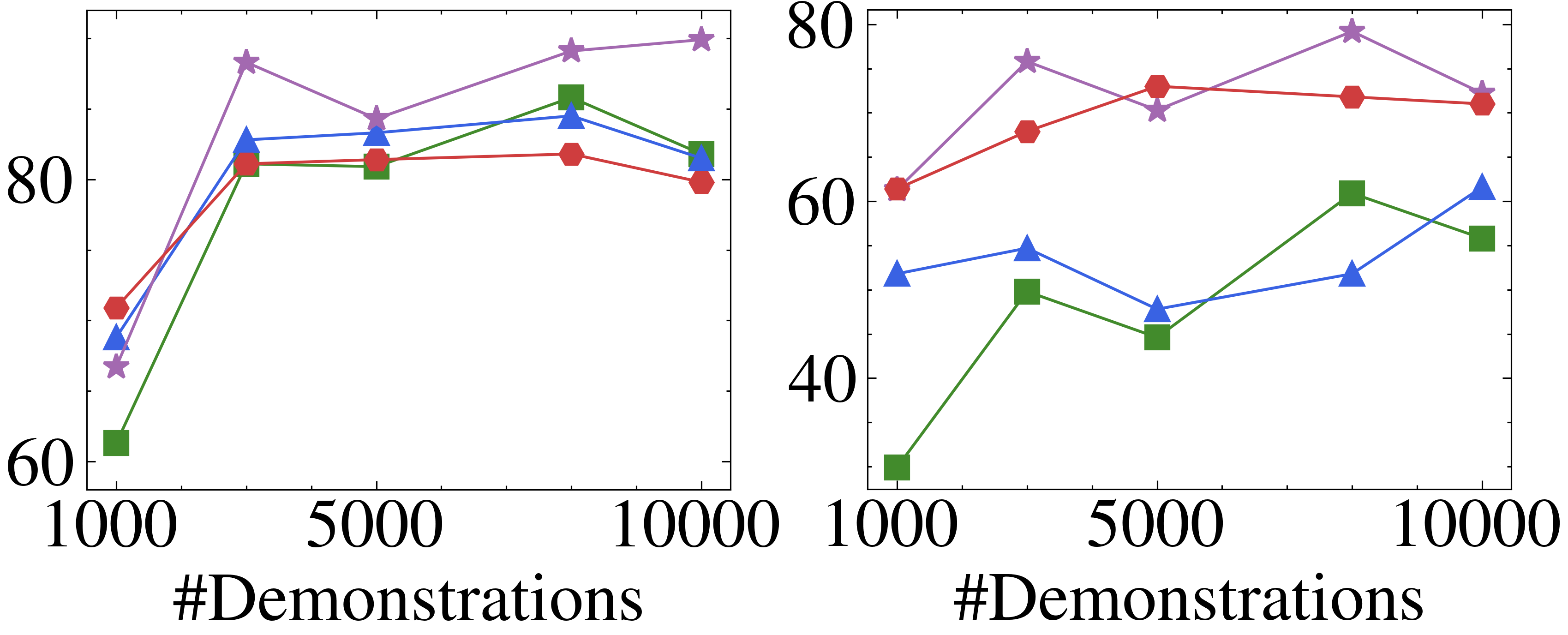}
  }
\vspace {-0.05in}
  \caption{Results under varying data budgets.}
\vspace {-0.1in}
  \label{varying_data}
\end{figure*}
\begin{table*}[t]
\caption{Results on Zero-Shot Generalization tasks.}
\vspace {-0.1in}
\label{tab:zero-shot}
\begin{center}
\begin{small}
\setlength{\tabcolsep}{1.6mm}{\begin{tabular}{l|cc|c|c|cc|c|c}
\toprule
Domain &\multicolumn{3}{c|}{BabyAI} &  \multicolumn{3}{c|}{Mini-BEHAVIOR}&  \multicolumn{2}{c}{Robotic Manipulation} \\ 
\midrule
\multirow{2}*{}& \multicolumn{2}{c|}{Train} & Eval & Train&
 \multicolumn{2}{|c|}{Eval}& Train&
 Eval \\
Task& Pickup & Open & Unlock & Throw 1 &Throw 2&Throw 3 & Putting-blocks &Novel combination\\
\midrule
BC    & 0.760$\ \pm\ $0.056& 0.983$\ \pm\ $0.021&0.120$\ \pm\ $0.010& 0.703$\ \pm\ $0.085&0.117$\ \pm\ $0.070&0.053$\ \pm\ $0.045
&0.597$\ \pm\ $0.032& 0.537$\ \pm\ $0.095 \\
DT & 0.783$\ \pm\ $0.031& 0.957$\ \pm\ $0.031& 0.057$\ \pm\ $0.051& 0.770$\ \pm\ $0.026&0.182$\ \pm\ $0.008&0.056$\ \pm\ $0.003
&0.550$\ \pm\ $0.019 &0.424$\ \pm\ $0.008\\
PDSketch   & \textbf{0.970$\ \pm\ $0.010}& 0.990$\ \pm\ $0.010&   0.127$\ \pm\ $0.021 & 0.013$\ \pm\ $0.006&> 5 minutes&> 5 minutes &- &-\\
ABIL-BC   & 0.937$\ \pm\ $0.021& \textbf{1.00} &0.980$\ \pm\ $0.026&0.717$\ \pm\ $0.055&0.590$\ \pm\ $0.013&0.485$\ \pm\ $0.054
&\textbf{0.860$\ \pm\ $0.027}&\textbf{0.857$\ \pm\ $0.044}\\
ABIL-DT    & 0.925$\ \pm\ $0.007& \textbf{1.00}& \textbf{0.993$\ \pm\ $0.012}&\textbf{0.783$\ \pm\ $0.031}&\textbf{0.702$\ \pm\ $0.034}&\textbf{0.585$\ \pm\ $0.120}
&0.822$\ \pm\ $0.026&0.823$\ \pm\ $0.050\\
\bottomrule
\end{tabular}}
\end{small}
\end{center}
\vskip -0.1in
\end{table*}

\begin{table}[t]
  \caption{Details of CLIPort benchmark.}
  \vspace{-2mm}
  \setlength{\tabcolsep}{1mm}{\begin{tabular}{lcc}
  \toprule
  Task  &Ave. Length & Evaluation \\
\midrule
 Packing-5shapes& 1 & 4/7 unseen/total colors \\
 Packing-20shapes & 1 & 4/7 unseen/total colors\\
 Placing-red-in-green & 2 & 11 total colors \\
 Putting-blocks-in-bowls & 2 & 7 total colors\\
 Assembling-kits& 5 & 10/5 total shapes/colors\\
 Separating-20piles& 7 & 7 total colors\\
  \bottomrule
  \end{tabular}\label{length:cliport}}
  \end{table}
  
\textbf{Results and Analysis.} 
The results on Mini-BEHAVIOR are provided in Table~\ref{result_mini}. 
In some simple short-horizon tasks, such as \textit{Installing a printer} and \textit{Opening packages}, BC shows satisfactory performance. 
Nevertheless, as the desired decision sequence grows, errors made by BC gradually accumulate, leading to an increasing deviation from the correct solution. 
This becomes particularly critical in the presence of disturbances or interferences, leading to a significant degradation on the generalization evaluation. 
Compared to BC, DT has achieved better performance in most tasks, however, it still performs poorly in generalization test, pointing out its vulnerability to environmental changes. 
In this benchmark, PDSketch failed to finish most tasks in the given time budget, due to the significant search depth required to achieve the goal. 
This highlights the limitations of model-based planning methods in long-horizon scenarios. Ensuring the learned transition remains accurate after numerous decision steps is challenging, making it difficult to provide a successful termination signal for the search. 
In contrast, ABIL performs reasoning at the symbolic level. Even if cleaning a car requires about 45 decision steps to complete, from the perspective of abstract operations, we can understand that we need to put the rag and soap back into the buckets after using them. This is similar to human's behavior, where we first recognize what the logical goal to be completed is, and then achieve it step by step through actions, rather than considering the impact of each limb movement, which would make the entire reasoning planning path too long. 
In this way, our neuro-symbolic ABIL successfully incorporates logic-based reasoning into imitation learning, achieving competitive results and showing good generalization under environmental change. 

\textbf{Further Analysis with Varying Horizons.}
As we discussed above, the generalization of imitation learning methods is closely related to the length of the task's horizon, especially in long-horizon tasks, where performance degradation is prone to occur. 
Mini-BEHAVIOR, which contains tasks that require different decision steps to complete, provides an appropriate observation window from this perspective. As shown in Table 2, the number of expert demonstrations required for these tasks ranges from 10 to 106 steps. 
On the one hand, we find that the increase in decision-making length required by the task indeed makes it more challenging, leading to a decline in the performance of almost all methods. On the other hand, we observe that the increase in decision-making length 
also amplifies performance differences in the generalization evaluation of the baseline methods compared to the basic evaluation. Our method demonstrates good generalization performance, aligning with the basic evaluation, and showcasing the potential of the ABIL in open environments. 

\subsection{Evaluation on Robotic Manipulation}
We further evaluate the proposal on the CLIPort~\cite{cliport} with 3D robotic manipulation tasks. 
In this environment, an agent needs to learn how to transport some objects and solve complex manipulation tasks based on visual observation. Every manipulation task needs to be achieved via a two-step primitive where each action involves a start and end-effector pose. 
Table~\ref{length:cliport} provides 
the average length of expert demonstrations. 
This benchmark involves the agent manipulating objects of various colors and shapes, reflecting the requirements in the open world, providing a greater challenge for imitation learning. 
Following~\cite{PDSketch, DBLP:conf/nips/HsuMTW23}, we represent each object with its image cropped from the observation with its pose, which could be completed by an external detection module. All demonstrations are collected using handcrafted oracle policies following CLIPort~\cite{cliport}, containing only successful trajectories. Since PDSketch mainly targets discrete actions, in this environment, we compared ABIL with BC and DT baselines.

\textbf{Results and Analysis.} The results are provided in Table~\ref{tab:cliport}. Although the execution length for robotic manipulation is shorter compared to household tasks in Mini-BEHAVIOR, the objects it needs to manipulate are more complex. As illustrated in Figure~\ref{ABIL_motivation}(b), in the packing-shapes task, the agent needs to manipulate objects of the same shape but unseen colors during testing. 
In packing-20shapes, which experts can complete in one step, pure learning-based BC and DT only achieved a 20\% success rate. However, our ABIL-BC achieved a satisfactory 94\% success rate through neuro-symbolic grounding to recognize the shape of corresponding objects. 
These results highlight the vulnerability of pure-learning-based methods in open-world scenarios and demonstrate the necessity of introducing neuro-symbolic reasoning in ABIL, which may provide a promising solution for household agents.

\begin{figure}[t]
  \centering
  \subfloat[Packing-20shapes]
  {
\label{cliport:grounding:subfig1}\includegraphics[width=0.48\linewidth]{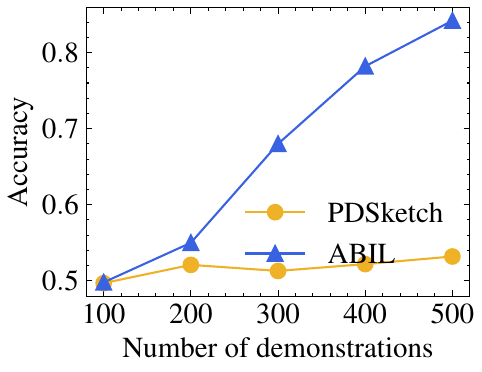}
  }
  \subfloat[Putting-blocks-in-bowls]
  {
\label{cliport:grounding:subfig2}\includegraphics[width=0.48\linewidth]{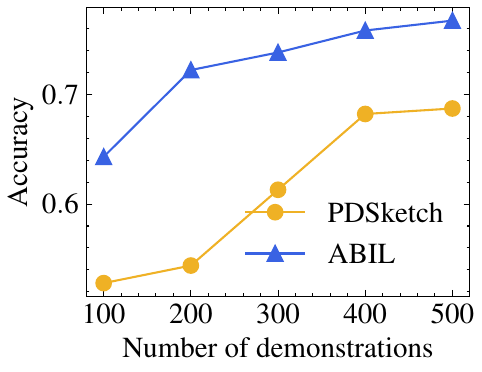}
  }
    \vspace{-2mm}
  \caption{Accuracy of neuro-symbolic grounding under varying data budgets in Robotic Manipulation tasks.}
\label{grounding_efficiency:cliport}
\end{figure}

\textbf{Comparison of Neuro-Symbolic Grounding.}
In robotic manipulation, the search based planning policy of PDSketch only applies to discrete symbolic operators, and therefore cannot be compared in our main experiments with continuous action space. We can only compare ABIL with PDSketch from the perspective of neuro-symbolic grounding. 
The results are provided in Figure~\ref{grounding_efficiency:cliport}.
Achieving precise grounding in this environment is more challenging. As the amount of demonstration increases, the accuracy of PDSketch rises slowly and erratically. In contrast, our method shows a rapid improvement, further demonstrating the advantage of our ABIL in terms of neuro-symbolic grounding efficiency. 

\subsection{Data Efficiency}

Data efficiency is important for imitation learning, especially in robotic tasks where expert demonstrations are usually expensive and scarce. 
In this subsection, we conduct experiments across varying sizes of expert demonstrations to evaluate the data efficiency. 

The results are provided in Figure~\ref{varying_data}. First, we found that PDSketch, the model-based planning method, achieved the best results on the \textit{pickup} task, even with a limited 500 demonstrations. However, as shown in Table~\ref{result_mini}, 
PDSketch fails in the complex long-horizon tasks. 
Second, we could find that in the simple task, \textit{pickup}, the generalization of different methods is consistently improved when the data volume increases. However, in the \textit{Opening packages} and \textit{Putting away dishes} tasks from Mini-BEHAVIOR,
although the results in basic evaluation improve with the increase of data volume, their performance in the corresponding generalization tests no longer grows. This also reflects the weakness of pure-learning-based methods, that is, they easily overfit to the specific training observations, and their performance in out-of-distribution observation is fragile.
Third, we found that our Abductive Imitation Learning framework has clearly improved the data efficiency of the BC and DT baselines, especially achieved significant generalization improvement in the out-of-distribution evaluation. 

\subsection{Zero-Shot Generalization}

Symbolic reasoning excels at generalization, especially ensuring the correctness of reasoning for any combination of logical clauses. 
In this subsection, we evaluate the zero-shot generalization performance in the composed tasks. 
In the BabyAI domain, we train the policies on the \textit{pickup} and \textit{open} task, then test them on the composed task \textit{unlock}. 
During training, the demonstrations from two tasks are mixed and learning in a multi-task scheme. 
In the Mini-BEHAVIOR domain, we primarily concentrate on generalization with the longer series of events, which demands the agent to make use of learned techniques for repeatedly completing a single task to achieve the desired goal. 
Take the \textit{Throwing away leftovers} task as an example, we train every model in the environment with 1 leftover hamburger to throw, while in the test environment, the agent is required to throw 2 or 3 hamburgers.
In robotic manipulation, we primarily focus on compositional generalization with the novel combination of goals, which demands the agent to re-combine learned concepts to achieve, as shown in Figure~\ref{ABIL_motivation}(c). 

The results are provided in Table~\ref{tab:zero-shot}. 
Although all baselines achieve satisfactory performance on the training tasks (\textit{pickup} and \textit{open}), their performance degrades on the simple combined task (\textit{unlock}). 
The pure-learning-based methods directly learn the action corresponding to the observations, lacking reasoning ability, thus unable to realize the need to first pick up a key that can open the target door, resulting in failure. 
PDSketch has reasoning ability, but its model-based planning solution accumulates errors with the increasing length of the sequence, resulting in poor performance and high computational overhead. In tasks with longer sequences, such as throwing away leftovers, solutions cannot be found even after running out of time. 
Our ABIL not only performs high-level reasoning to know that sub-goals should be sequentially completed but also can zero-shot achieve the composed tasks. 

\section{Conclusion and Future Work}

In this work, we proposed a novel framework, \textbf{AB}ductive \textbf{I}mitation \textbf{L}earning (ABIL), which integrates data-driven learning with symbolic reasoning to address long-horizon tasks in imitation learning. 
ABIL bridges the gap between neural perception and logical reasoning by autonomously generating predicate candidates from raw observations with the knowledge base, enabling effective reasoning without requiring extensive manual annotations. Experiments demonstrate that ABIL significantly improves data efficiency and generalization across various long-horizon tasks, positioning it as a promising neuro-symbolic solution for imitation learning. 

Despite its contributions, ABIL has several limitations that suggest promising directions for future work: (1) Uncertainty and partial observability: The current framework assumes deterministic and fully observable environments, consistent with existing work~\cite{PDSketch, DBLP:conf/corl/SilverATLK22, DBLP:conf/aaai/SilverCKMLKT23}. However, real-world environments are often stochastic and partially observable. A promising direction is to explore POMDP techniques~\cite{DBLP:conf/uai/GangwaniL0019, DBLP:conf/icra/GarrettPLKF20}, which would allow ABIL to maintain a belief space and sample actions under uncertainty. 
(2) Automatic knowledge learning: 
Like most neuro-symbolic and abductive learning work, ABIL assumes the availability of a symbolic solution and relies on an accurate and sufficient knowledge base~\cite{ABL_DaiX0Z19, PDSketch}. 
A key direction is to incorporate advanced knowledge learning techniques~\cite{DBLP:conf/iclr/WangMHZ0G23, DBLP:conf/aaai/YangSTLDZ24} to reduce reliance on human-defined knowledge. Additionally, introducing the active learning manner~\cite{DBLP:journals/corr/abs-2407-15786} with human feedback could help correct and supplement the knowledge base, further enhancing ABIL’s adaptability and robustness. 
In summary, ABIL offers a timely and promising solution for neuro-symbolic imitation learning, particularly for long-horizon planning. Addressing the challenges of uncertain environments and incomplete knowledge will unlock its full potential, making it a reliable system for real-world applications.

\begin{acks}
This research was supported by Leading-edge Technology Program of Jiangsu Science Foundation (BK20232003), Key Program of Jiangsu Science Foundation (BK20243012) and the Postgraduate Research \&
Practice Innovation Program of Jiangsu Province (KYCX24\_0233). 
\end{acks}

\bibliographystyle{ACM-Reference-Format}
\bibliography{kdd25}

\appendix



\begin{table*}[t]
\centering
\caption{Results under varying gorunding accuracy.}
  \setlength{\tabcolsep}{2mm}{\begin{tabular}{lllcccccc}
  \toprule
 \multicolumn{3}{c}{Grounding accuracy}& 100\% & 90\% &75\%& 50\% & Data-driven  \\
  \midrule
\multirow{3}*{ABIL-BC} &\multirow{2}*{Pickup}&Basic  & 0.847$\ \pm\ $0.025 $\uparrow$&	0.787$\ \pm\ $0.015 $\uparrow$&0.763$\ \pm\ $0.023 $\uparrow$&	0.717$\ \pm\ $0.021&	0.723$\ \pm\ $0.031 (BC)\\
&&Gen  & 0.730$\ \pm\ $0.010 $\uparrow$&	0.667$\ \pm\ $0.057 $\uparrow$&0.653$\ \pm\ $0.023 $\uparrow$&	0.560$\ \pm\ $0.017 $\uparrow$&	0.533$\ \pm\ $0.031 (BC)\\
\cmidrule{2-8}
& \multicolumn{2}{l}{Putting-blocks-in-bowls}  & - &	0.962$\ \pm\ $0.012 $\uparrow$& 0.599$\ \pm\ $0.036 $\uparrow$ &0.512$\ \pm\ $0.022 $\uparrow$&	0.507$\ \pm\ $0.030 (BC)\\
\midrule
\multirow{3}*{ABIL-DT} &\multirow{2}*{Pickup}&Basic&0.845$\ \pm\ $0.035 $\uparrow$ & 0.860$\ \pm\ $0.017 $\uparrow$&	0.800$\ \pm\ $0.036 $\uparrow$& 0.733$\ \pm\ $0.025 $\uparrow$	&0.490$\ \pm\ $0.040 (DT)\\
&&Gen&0.763$\ \pm\ $0.051 $\uparrow$ & 0.740$\ \pm\ $0.056 $\uparrow$&	0.597$\ \pm\ $0.042 $\uparrow$& 0.563$\ \pm\ $0.029 $\uparrow$	&0.320$\ \pm\ $0.070 (DT)\\
\cmidrule{2-8}
& \multicolumn{2}{l}{Putting-blocks-in-bowls}  & - &	0.917$\ \pm\ $0.033 $\uparrow$& 0.576$\ \pm\ $0.041 $\uparrow$&	0.462$\ \pm\ $0.047&	0.539$\ \pm\ $0.068 (DT)\\
  \bottomrule
  \end{tabular}\label{tab:grouding_error}}
\end{table*}
\begin{table*}[t]
\centering
\caption{Additional Results on Zero-Shot Generalization tasks.}
  \setlength{\tabcolsep}{2mm}{\begin{tabular}{llc|ccccc}
  \toprule
Domain&  &Task& BC & DT &PDSketch& ABIL-BC & ABIL-DT  \\
  \midrule
  \multirow{3}*{} &Train& Opening 1 package& 0.950$\ \pm\ $0.087& \textbf{1.00} &0.467$\ \pm\ $0.057 &0.997$\ \pm\ $0.006& \textbf{1.00} \\
  \cmidrule{2-8}
 Mini-BEHAVIOR &\multirow{2}*{Eval}  &Opening 2 packages& 0.012$\ \pm\ $0.010& 0.037$\ \pm\ $0.025 &0.020$\ \pm\ $0.010 & 0.818$\ \pm\ $0.014 & \textbf{0.840$\ \pm\ $0.035}\\
  & &Opening 3 packages& 0.002$\ \pm\ $0.004& 0.024$\ \pm\ $0.008 &> 5 minutes& 0.551$\ \pm\ $0.032& \textbf{0.631$\ \pm\ $0.041}\\
  \bottomrule
  \end{tabular}\label{tab:more-zero-shot}}
\end{table*}
\section{Experimental Details}
\subsection{BabyAI and Mini-BEHAVIOR}
In these two environments, each object feature is represented by its state and position in the room, and the robot feature is represented by its position and direction. The state representation is composed of features of all objects and the robot. The action spaces are both discrete. All expert demonstrations are generated by scripts based on $A^*$ search. 
In BabyAI, 1000 demonstrations were used for training per task and to obtain Table~\ref{tab:babyai}. In Mini-BEHAVIOR, \textit{install-a-printer}, \textit{opening packages}, and \textit{moving boxes to storage} used 1000, while other tasks used 3000 expert demonstrations for training. 

\textbf{Model Architecture.} For each predicate (e.g. is-dusty), we build a binary classifier, which takes a single object $o$ as argument, and returns a scalar value from 0 to 1, indicating the classification score. All methods use the same network architecture which is based on the Neural Logic Machine~\cite{NLM}. Specifically, we first encode the state with a two-layer NLM. For BC, we use a single linear layer, taking the state embedding as the input and output actions. For DT, we build a single transformer layer following the two-layer encoder, with the causal mask to generate future action with past states and actions. For PDSketch, we choose the full mode in the original paper~\cite{PDSketch}. For ABIL-BC, we implement the behavior modules using BC model, and for ABIL-DT, we implement the behavior modules using DT model.

\subsection{Robotic Manipulation}
\label{add_RM}
In this environment, each object is represented as a tuple of a 3D xyz location, and an image crop. Following~\cite{PDSketch}, we first compute the 2D bounding box of the object in the camera plane, then crop the image patch and resize it to 24 by 24 to obtain the image crop. The action space is continuous, each action involves a start and end-effector pose. 
Table~\ref{length:cliport} provides 
the average length of expert demonstrations. 
This benchmark involves the agent manipulating objects of various colors and shapes, reflecting the requirements in the open world, providing a greater challenge for imitation learning. 
For each task, 1000 expert demonstrations were used for training and to obtain Table~\ref{tab:cliport}. All of these demonstrations are collected using oracle policies following CLIPort~\cite{cliport}, containing only successful trajectories. 
  
\textbf{Model Architecture.} Image feature of each object is a 64-dimensional embedding obtained via an image encoder, which is a 3-layer convolutional neural network followed by a linear transformation layer. For each predicate (e.g. is-red), we build a binary classifier, which takes the image feature of an object, and returns a scalar value from 0 to 1, indicating the classification score. The model implementation is same as in BabyAI and Mini-BEHAVIOR, except output continuous value as action.

\section{Supplemental Results}  

\subsection{Study on Performance with Imperfect Symbolic Grounding}

To evaluate the influences of neuro-symbolic errors upon ABIL, we further conduct experiments on the \textit{Pickup} and \textit{Putting-blocks-in-bowls} task . Experimental results are provided in Table~\ref{tab:grouding_error}.

Like human reasoning, incorrect logical objectives may lead to the failure of sequential decision-making. Neuro-symbolic errors indeed lower the performance of ABIL. Nevertheless, ABIL integrates data-driven imitation and logical objectives in learning, it has a tolerance for neuro-symbolic errors. Even under 75\% grounding accuracy, it can still achieve improvements compared to purely data-driven methods. 

\subsection{Additional Results on zero-shot generalization}

In Table~\ref{tab:more-zero-shot}, we provide additional evaluation results on zero-shot generalization task in the Mini-BEHAVIOR benchmark. In \textit{Opening Packages} task, we train every model in the environment with 1 package to open, while in the testing environment, the agent is required to open 2 or 3 packages.

\section{Reproducibility}

To promote reproducibility, we release the code on GitHub\footnote{\url{https://github.com/Hoar012/ABIL-KDD-2025}}. This may also assist future research.

\section{Details of Knowledge Base}
\label{app_kb}

In this section, we provide a detailed illustration of our knowledge base to help readers understand and reproduce. 
\onecolumn
\subsection{BabyAI}

\begin{itemize}
\item Goto a red box
\begin{figure}[H]
\centering
  \includegraphics[width=0.3\linewidth]{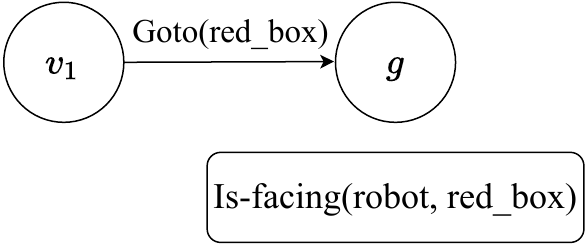}
\end{figure}
\item Pickup a red key
\begin{figure}[H]
\centering
  \includegraphics[width=0.5\linewidth]{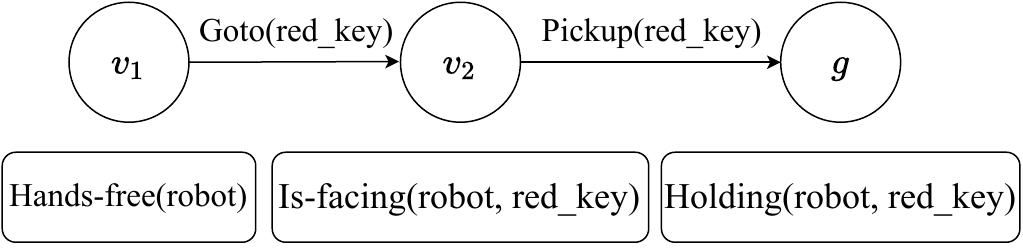}
\end{figure}
\item Open a red door
\begin{figure}[H]
\centering
  \includegraphics[width=0.5\linewidth]{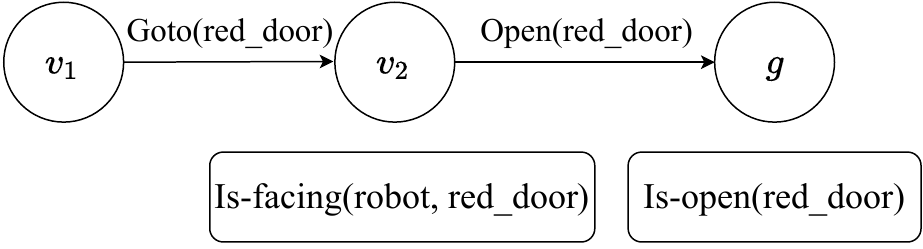}
\end{figure}
\item Put the ball next to the box
\begin{figure}[H]
\centering
  \includegraphics[width=0.7\linewidth]{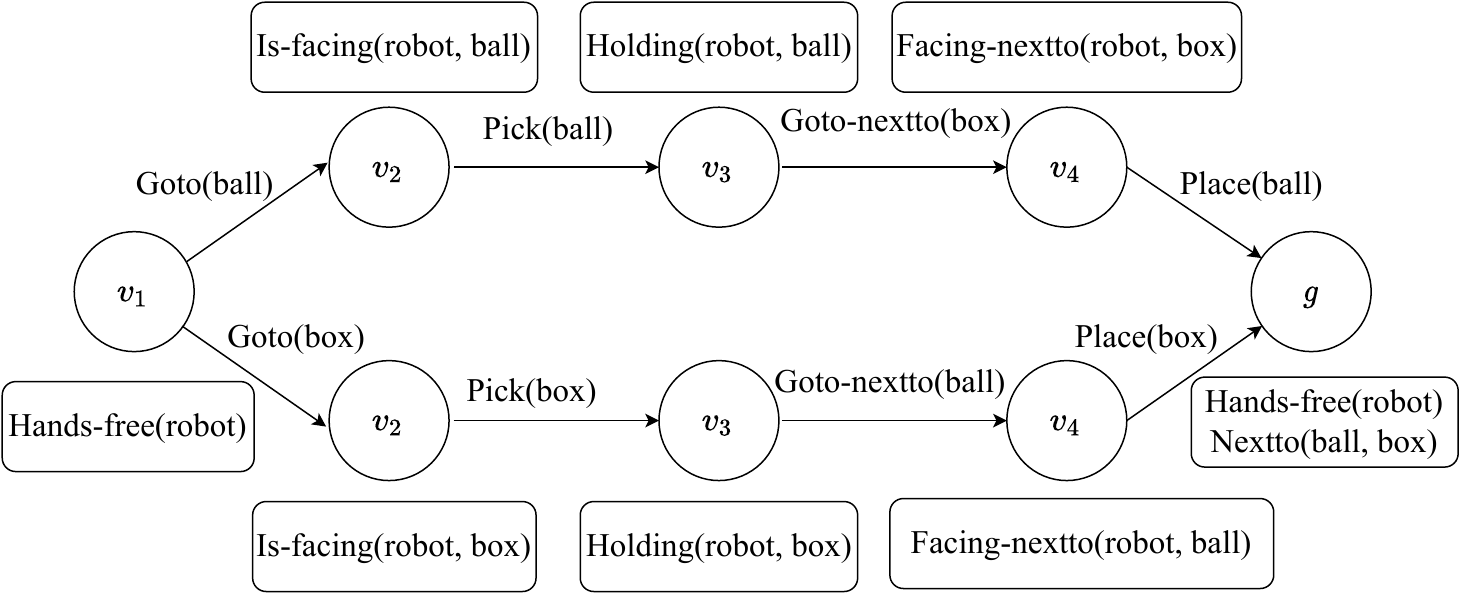}
\end{figure}
\item Unlock a red door
\begin{figure}[H]
\centering
  \includegraphics[width=0.75\linewidth]{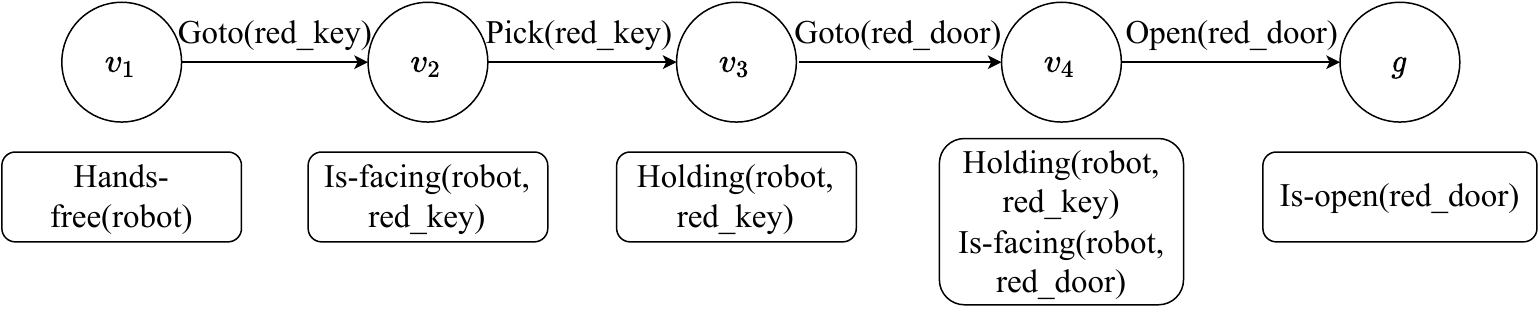}
\end{figure}
\end{itemize}

\subsection{Mini-BEHAVIOR}
In this domain, our state machine is mainly composed of several typical categories. For tasks mainly about tidying up the room, e.g. \textit{Throwing away leftovers}, we split the primitive actions into $op_{pick}$ and $op_{place}$, which is required to perform an action sequence to finish pickup or place subtask. Combined with our symbolic-grounding $f$, the agent will be able to distinguish when and where to pick and place. For tasks mainly about cleaning, e.g. \textit{Cleaning a car}, we split the primitive actions into $op_{clean}$ and $op_{put}$, which is required to finish washing or putting subtask. In addition, some tasks involve more operators, such as \textit{install a printer}.
We provide detailed illustrations of these representative state machine models.

\begin{itemize}
\item Throwing away leftovers
 
 In this task, there are 3 hamburgers on plates, which are on a countertop in the kitchen. The agent must throw all of the hamburgers into the ashcan. 
\begin{figure}[H]
\centering
  \includegraphics[width=0.9\linewidth]{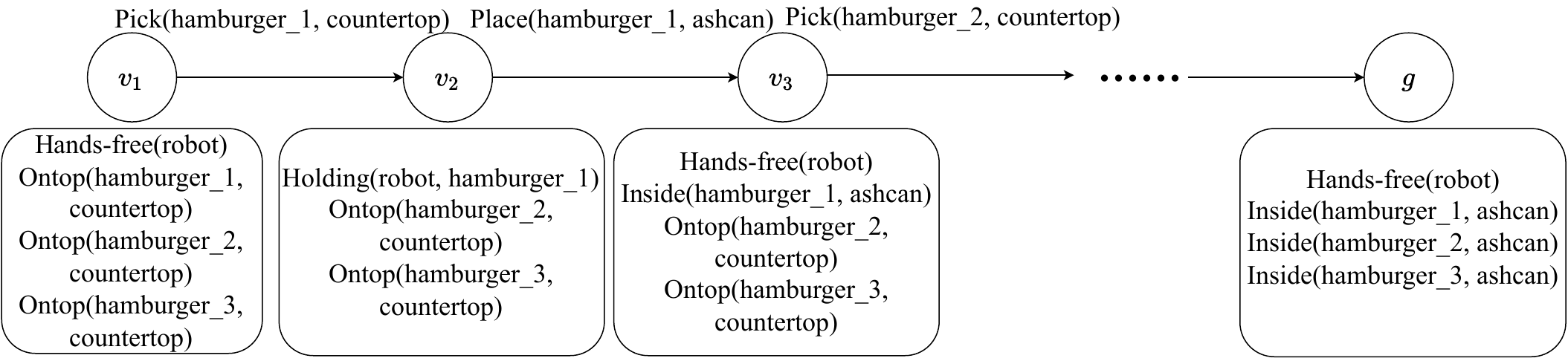}
\end{figure}
    \item Cleaning a car
 
 In this task, initially there is a dusty car, a soap and unsoaked rag, the agent need to use the rag and soap to clean the car. Finally the agent should place the rag and soap in a bucket.
\begin{figure}[H]
\centering
  \includegraphics[width=0.6\linewidth]{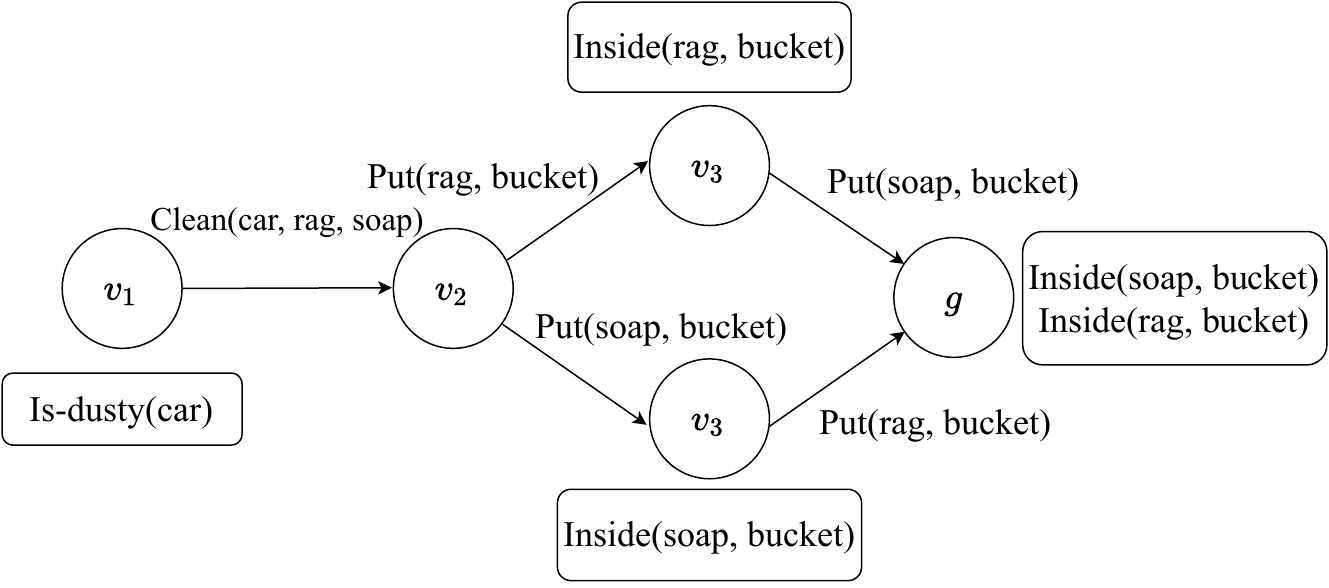}
\end{figure}
 
\item Installing a printer
 
 In this task, initially there is a printer on the floor, and the agent must place it on the table and toggle it on.
 \begin{figure}[H]
 \centering
  \includegraphics[width=0.8\linewidth]{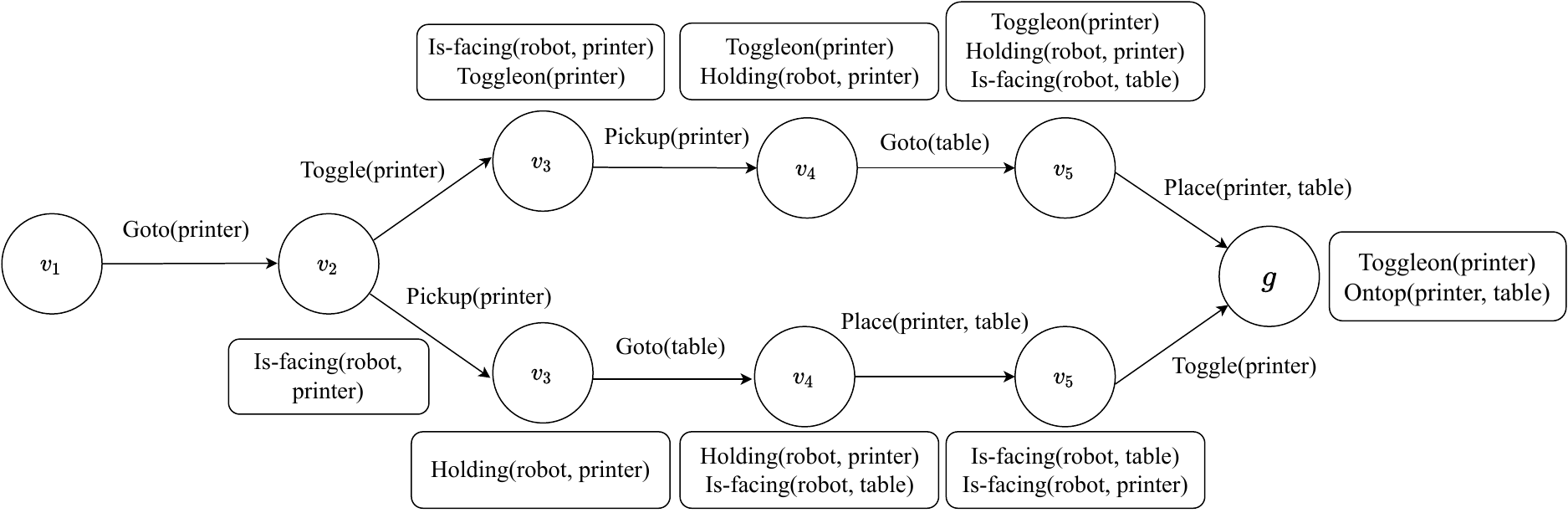}
\end{figure}

\end{itemize}

\subsection{Robotic Munipulation}
\begin{itemize}
    \item Packing star into the box
\begin{figure}[H]
\centering
\vspace{-0.3in}
  \includegraphics[width=0.28\linewidth]{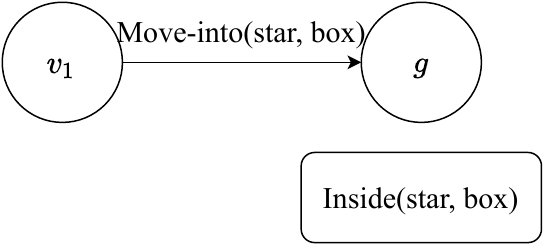}
\end{figure}
\vspace{-0.1in}
    \item Putting-red plocks-in-green bowls
\begin{figure}[H]
\centering
  \includegraphics[width=0.7\linewidth]{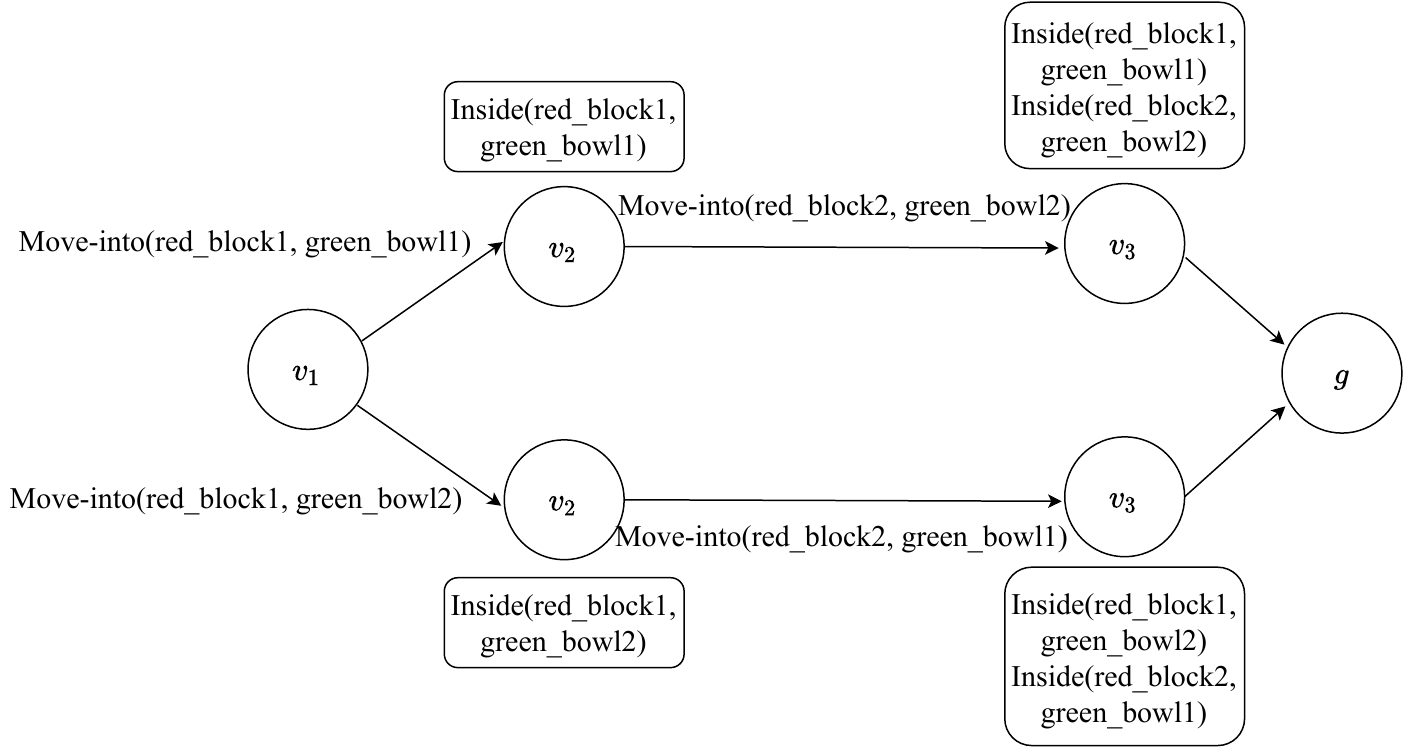}
\end{figure}
    \item Separating-piles
\begin{figure}[H]
\centering
  \includegraphics[width=0.3\linewidth]{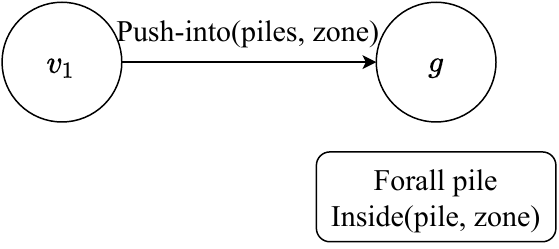}
\vspace{-0.1in}
\end{figure}
    \item Assembling-kits
    \begin{figure}[H]
    \centering
  \includegraphics[width=0.8\linewidth]{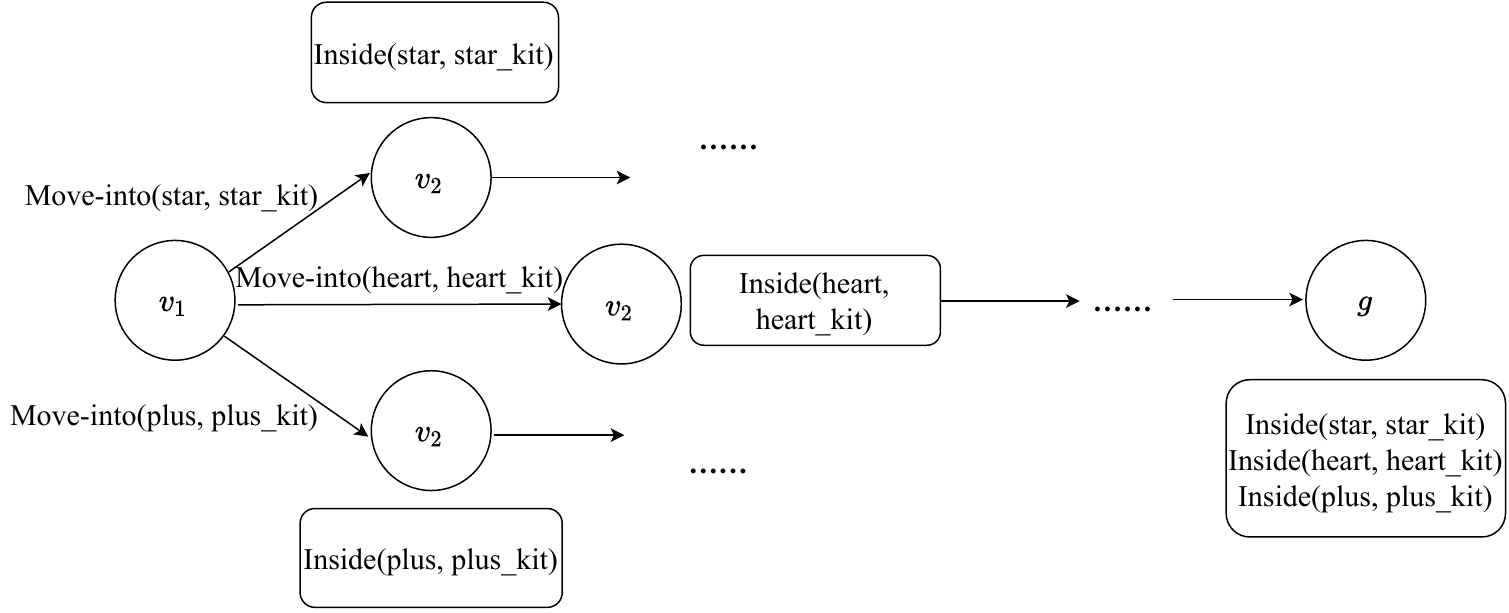}
\end{figure}
\end{itemize}

\end{document}